\documentclass{article} % For LaTeX2e
\usepackage{iclr2026_conference,times}

\usepackage{amsmath,amsfonts,bm}

\def\eqref#1{equation~\ref{#1}}
\def\1{\bm{1}}

\DeclareMathAlphabet{\mathsfit}{\encodingdefault}{\sfdefault}{m}{sl}
\SetMathAlphabet{\mathsfit}{bold}{\encodingdefault}{\sfdefault}{bx}{n}

\usepackage{hyperref}
\usepackage{url}
\newcommand{\name}{ReCogDrive}

\usepackage[utf8]{inputenc} % allow utf-8 input
\usepackage[T1]{fontenc}    % use 8-bit T1 fonts
\usepackage{hyperref}       % hyperlinks
\usepackage{url}            % simple URL typesetting
\usepackage{booktabs}       % professional-quality tables
\usepackage{amsfonts}       % blackboard math symbols
\usepackage{nicefrac}       % compact symbols for 1/2, etc.
\usepackage{microtype}      % microtypography
\usepackage{xcolor}         % colors
\usepackage{acronym}
\usepackage{graphicx}
\usepackage{wrapfig}

\usepackage{amsmath}
\usepackage{amssymb}
\usepackage{mathtools}
\usepackage{amsthm}
\usepackage{algorithm}
\usepackage{algpseudocode}
\usepackage{setspace} % Include the setspace package
\usepackage{enumitem}
\usepackage{acronym}
\usepackage{xspace}

\usepackage{graphicx}    % 用于 \resizebox
\usepackage{array}       % 用于自定义表格列
\usepackage[misc]{ifsym} % letters icon
\usepackage{makecell}    % 用于 \makecell
\usepackage{booktabs}    % 用于 \toprule 和 \bottomrule
\usepackage{setspace}    % 用于 \setstretch

\usepackage{multirow}
\usepackage{indentfirst}
\usepackage{subcaption}
\usepackage{xcolor}
\usepackage[table]{xcolor}
\usepackage{pifont}
\usepackage{lineno}

\usepackage[capitalize,noabbrev]{cleveref}

\theoremstyle{plain}

\theoremstyle{definition}

\theoremstyle{remark}

\acrodef{vla}[VLA]{Vision-Language-Action}
\acrodef{llms}[LLMs]{Large Language Models}
\acrodef{vlms}[VLMs]{Vision-Language Models}
\acrodef{vae}[VAE]{variational autoencoder}
\acrodef{cvae}[CVAE]{conditional variational autoencoder}
\usepackage[textsize=tiny]{todonotes}

\usepackage[capitalize]{cleveref}

\crefname{algorithm}{Alg.}{Algs.}
\Crefname{algocf}{Algorithm}{Algorithms}
\crefname{section}{Sec.}{Secs.}
\Crefname{section}{Section}{Sections}
\crefname{table}{Tab.}{Tabs.}
\Crefname{table}{Table}{Tables}
\crefname{figure}{Fig.}{Figs.}
\Crefname{figure}{Figure}{Figures}
\crefname{equation}{Eq.}{Eqs.}
\Crefname{equation}{Equation}{Equations}
\crefname{appendix}{Appx.}{Appxs.}
\Crefname{appendix}{Appendix}{Appendices}
\crefformat{equation}{Eq.~#2\textup{#1}#3}

\definecolor{darkgreen}{rgb}{0.09, 0.45, 0.27}
\definecolor{alizarin}{rgb}{0.82, 0.1, 0.26}
\definecolor{new_cyan}{rgb}{0.10, 0.62, 0.57}

\definecolor{robo_blue}{RGB}{66, 133, 244}
\definecolor{robo_red}{RGB}{192, 0, 0}
\definecolor{robo_yellow}{RGB}{251, 189, 5}
\definecolor{robo_green}{RGB}{0, 176, 80}
\definecolor{robo_gray}{RGB}{100, 100, 100}
\definecolor{customblue}{HTML}{005AD7}

\title{ReCogDrive: A Reinforced Cognitive Framework for End-to-End Autonomous Driving}
\author{
    \textbf{Yongkang Li}\textsuperscript{1,2 *}, 
    \textbf{Kaixin Xiong}\textsuperscript{2 *}, 
    \textbf{Xiangyu Guo}\textsuperscript{1,2},
    \textbf{Fang Li}\textsuperscript{2},
    \textbf{Sixu Yan}\textsuperscript{1},
    \textbf{Gangwei Xu}\textsuperscript{1,2}, \\
    \textbf{Lijun Zhou}\textsuperscript{2}\textbf{,}
    \textbf{Long Chen}\textsuperscript{2}\textbf{,}
    \textbf{Haiyang Sun}\textsuperscript{2 $\dagger$}\textbf{,}
    \textbf{Bing Wang}\textsuperscript{2}\textbf{,}
    \textbf{Kun Ma}\textsuperscript{2}\textbf{,}
    \textbf{Guang Chen}\textsuperscript{2}\textbf{,} \\
    \textbf{Hangjun Ye}\textsuperscript{2}\textbf{,}
    \textbf{Wenyu Liu}\textsuperscript{1}\textbf{,}
    \textbf{Xinggang Wang}\textsuperscript{1}~\textsuperscript{\Letter} \\
    \\
    \textsuperscript{1}Huazhong University of Science and Technology \hspace{1em}
    \textsuperscript{2}Xiaomi EV \\
    {\tt\small \{liyk, xgwang\}@hust.edu.cn, \{sunhaiyang1\}@xiaomi.com} \\
    {{\tt\small \textbf{Model \& Data}: \href{https://huggingface.co/collections/owl10/recogdrive-68bafa143de172bab8de5752}{\textcolor{customblue}{ReCogDrive HuggingFace Collection}}}} \\
  {{\tt\small \textbf{Code}: \href{https://github.com/xiaomi-research/recogdrive}{\textcolor{customblue}{ReCogDrive Github Repository}}}} 
}
\iclrfinalcopy 
\begin{document}

\maketitle

\begin{abstract}

Recent studies have explored leveraging the world knowledge and cognitive capabilities of \ac{vlms} to address the long-tail problem in end-to-end autonomous driving. However, existing methods typically formulate trajectory planning as a language modeling task, where physical actions are output in the language space, potentially leading to issues such as format-violating outputs, infeasible actions, and slow inference speeds. In this paper, we propose ReCogDrive, a novel \textbf{Re}inforced \textbf{Cog}nitive framework for end-to-end autonomous \textbf{Driv}ing, unifying driving understanding and planning by integrating an autoregressive model with a diffusion planner. First, to instill human driving cognition into the VLM, we introduce a hierarchical data pipeline that mimics the sequential cognitive process of human drivers through three stages: generation, refinement, and quality control. Building on this cognitive foundation, we then address the language-action mismatch by injecting the VLM's learned driving priors into a diffusion planner to efficiently generate continuous and stable trajectories. Furthermore, to enhance driving safety and reduce collisions, we introduce a Diffusion Group Relative Policy Optimization (DiffGRPO) stage, reinforcing the planner for enhanced safety and comfort. Extensive experiments on the NAVSIM and Bench2Drive benchmarks demonstrate that ReCogDrive achieves state-of-the-art performance. Additionally, qualitative results across diverse driving scenarios and DriveBench highlight the model's scene comprehension. Code and models are available at \href{https://github.com/xiaomi-research/recogdrive}{https://github.com/xiaomi-research/recogdrive}.

% ReCogDrive enables the description of driving scenes and trajectory planning, utilizing multi-view camera images alongside driving instructions

%第二三句合并，合并说
%缺陷两点
%propose方法回扣下缺陷
% 1. ..., task-aware semantic reasoning and trajectory generation with long-horizon (spatio-temporal) consistency remain challenge.
% 2. limitations of existing methods: (1)...; (2)...
% 3. In this work, we propose ... (1) (2) (3). explain how to solve abovementioned limitations
% 4. experiment results
% 5. highlight your contribution in your domain. (should be mentioned at the end of the introduction) Our evaluations underscore the potential of VLMs to enhance the generalization of autonomous driving, while also highlighting the critical role of ... (GRPO).
\end{abstract}    
\section{Introduction}
\label{sec:intro}

Autonomous driving, which aims to predict a smooth, comfortable, and collision-free trajectory for a vehicle, has seen significant advancements. Recent end-to-end autonomous driving systems~\citep{jiang2023vad,hu2023planning,chen2024vadv2,hu2022st, zhang2024sparsead} unify the perception~\citep{jiang2024far3d,philion2020lift,zhang2023fully}, prediction~\citep{chai2019multipath,gu2023vip3d}, and planning~\citep{chitta2022transfuser,liao2024diffusiondrive} modules into a single pipeline for joint optimization, demonstrating impressive performance under open-loop evaluation. However, these systems often fail to generalize to long-tail scenarios, where data is limited and the driving conditions deviate significantly from those encountered during training.

Recent research~\citep{tian2024drivevlm,jiang2024senna,hwang2024emma} addresses the long-tail challenge by introducing \ac{vlms}, which are pre-trained on large-scale internet datasets and exhibit strong generalization abilities along with rich world knowledge. Specifically, \ac{vlms} applications in autonomous driving can be categorized into two main approaches: (1) \textit{dual-system} approaches~\citep{jiang2024senna,tian2024drivevlm}, where \ac{vlms} generate low-frequency trajectories or high-level commands to guide an end-to-end driving system; and (2) \textit{single-system} approaches~\citep{hwang2024emma,mao2023gpt,xing2024openemma,mao2023language,wang2024omnidrive,bai20243d,zhang2024wisead,zhao2025sce2drivex}, where standalone \ac{vlms} directly predict future trajectories and can be optimized in an end-to-end manner. Most of these approaches reformulate the motion planning task as a language modeling problem, generating trajectories in a textual format through autoregressive prediction. These methods leverage the generalization of the models and chain-of-thought~\citep{wei2022chain} reasoning to enhance interpretability and reasoning in complex scenarios. 

Despite these advantages, directly applying pre-trained VLMs by reformulating trajectory planning as a text generation task for autonomous driving reveals three critical limitations: (1) \textit{Domain discrepancy of pre-trained knowledge.} VLMs~\citep{mao2023gpt,mao2023language} are trained on generic internet data that lacks the nuanced knowledge required for driving, resulting in a significant domain gap. (2) \textit{Modality mismatch in trajectory generation.} The discrete language space of VLMs fundamentally conflicts with the continuous action space required for planning. This mismatch leads to practical issues, as the probabilistic nature of autoregressive decoding can lead not only to physically infeasible trajectories but also to format-violating outputs that result in parsing errors. (3) \textit{Suboptimal policies from imitation learning.} The heavy reliance on behavior cloning causes models~\citep{hwang2024emma} to converge to a suboptimal policy that struggles in rare scenarios and can be unsafe.

\begin{figure}[t!]
    \centering
    \includegraphics[width=\linewidth]{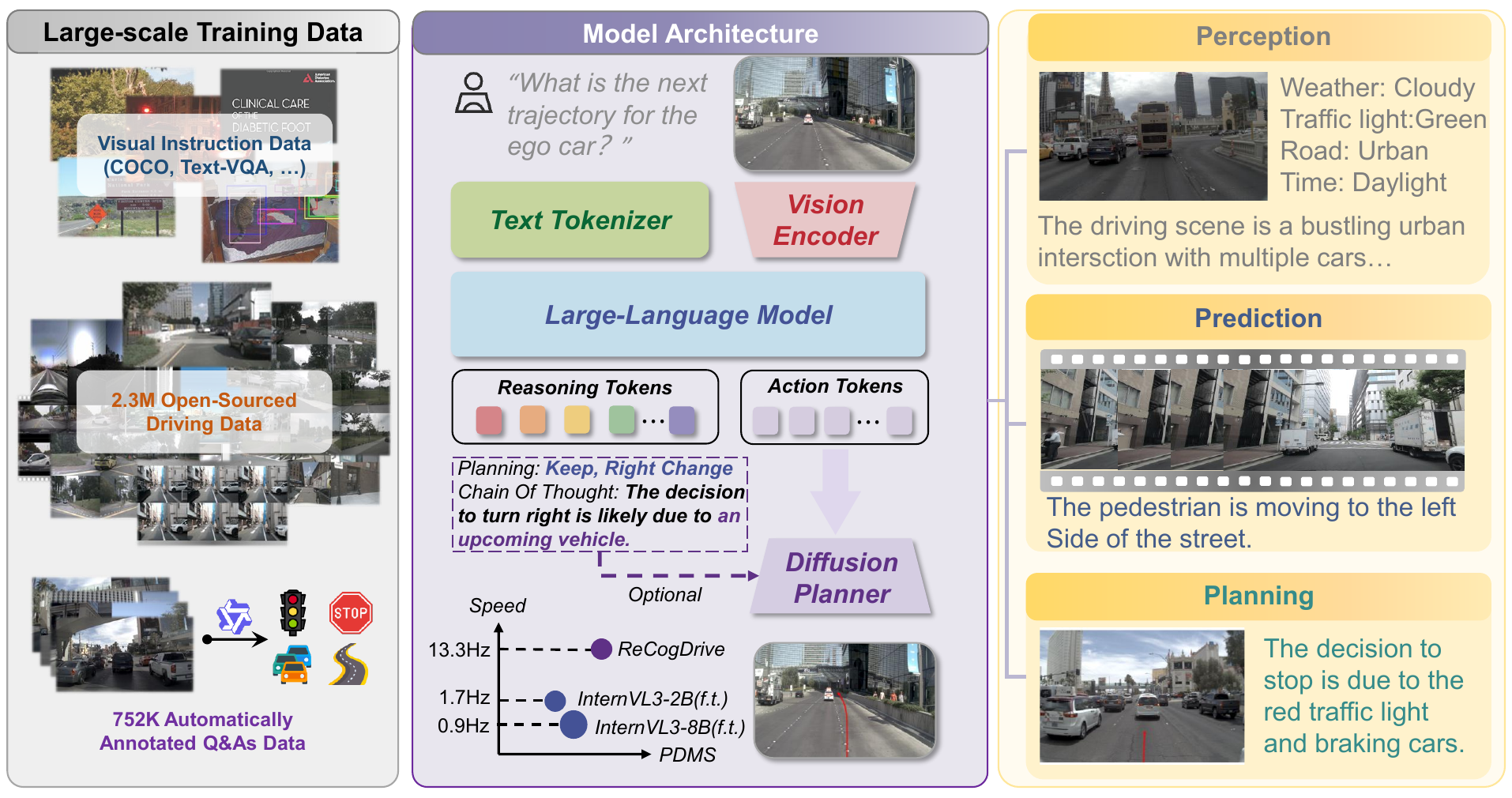}
    \caption{\textbf{Overview of \name{}.} We present \name{}, an end-to-end autonomous driving system, which possesses rich driving priors and generates continuous, stable trajectories via a diffusion denoising process. \name{} is capable of performing tasks spanning from low-level scene perception and motion prediction to high-level driving planning and decision making.}
    \label{fig:representation_gen}
    \vspace{-0.3cm}
\end{figure}

To address these challenges, we propose \name{}, a novel end-to-end autonomous driving system that integrates the cognitive reasoning of Vision-Language Models (VLMs) with a reinforcement learning-enhanced diffusion planner. Our framework introduces three key innovations: First, to bridge the domain gap, we design a hierarchical data pipeline that, through three stages of generation, refinement, and quality control, constructs a large-scale VQA dataset to instill human-like cognitive priors into the VLM. Second, to resolve the modality mismatch, we design a diffusion planner that effectively translates the VLM's latent cognitive representations into continuous and stable driving trajectories. Finally, to overcome the limitations of imitation learning, we introduce a reinforcement learning stage based on Diffusion Group Relative Policy Optimization (DiffGRPO), enabling the planner to explore safer, more optimal behaviors beyond the expert dataset.

% reinforcement learning allows the model to generate safe, comfortable, and stable trajectories through interaction with closed-loop evaluation signals.

We extensively evaluate ReCogDrive on NAVSIM and Bench2Drive, achieving new state-of-the-art performance in both open-loop and closed-loop settings. Additionally, we test ReCogDrive-VLM on the DriveLM and DriveBench benchmarks, demonstrating its superiority over both closed-source generalist models and open-source expert models. Qualitative analyses further demonstrate that our reinforcement learning method enables the model to drive more safely and comfortably while significantly reducing collisions. The main contributions of this work are as follows:
\begin{itemize}[leftmargin=2.0em]

\item We propose \name{}, a novel end-to-end autonomous driving framework that integrates a cognitive VLM with a diffusion planner, leveraging the rich world knowledge of VLMs to provide cognitive guidance and enabling stable and precise trajectory generation. 

\item We propose Diffusion Group Relative Policy Optimization (DiffGRPO) to fine-tune the planner, allowing the model to learn directly from experience and optimize for safety and comfort, moving beyond mere imitation.

\item We establish new state-of-the-art performance on NAVSIM and Bench2Drive in both open-loop and closed-loop evaluations, demonstrating significant improvements over prior approaches.
% We present a three-stage training framework: initially adapting the VLM to driving scenarios using a large-scale driving question-answering dataset, subsequently training the diffusion policy model to generate safe, comfortable, and reliable trajectories through imitation learning, and finally enhancing performance via reinforcement learning.
%\color{red}{We outline a three-stage training framework: initially adapting the VLM to driving scenarios using a large-scale driving question-answering dataset, subsequently training the diffusion model to generate safe, comfortable, and reliable trajectories through imitation learning, and finally enhancing performance via reinforcement learning.}

% We carry out extensive evaluations across NAVSIM benchmark. Our method attains a state-of-the-art PDMS score of 89.6 in closed-loop evaluation, underscoring its effectiveness and real-world viability.

\end{itemize}

\section{Related Work}
\label{sec:formatting}

\noindent\textbf{Vision-Language-Models in Autonomous Driving.}
Numerous studies have leveraged the world knowledge embedded in \ac{vlms} to explore their application in autonomous driving. Current approaches for autonomous driving using \ac{vlms} can be categorized primarily into two types: \textit{dual-system}~\citep{tian2024drivevlm,jiang2024senna} and \textit{single-system}~\citep{hwang2024emma,mao2023gpt,xing2024openemma,mao2023language,wang2024omnidrive,bai20243d,shao2024lmdrive,zhang2024wisead,zhao2025sce2drivex, fu2025orion}. Dual-system methods, such as DriveVLM and Senna, integrate \ac{vlms} with end-to-end driving systems, where \ac{vlms} generate low-frequency trajectories or high-level commands that guide the end-to-end model in producing the final trajectory. In contrast, single-system methods, such as GPT-Driver and EMMA, reformulate the trajectory prediction task as a language modeling problem and leverage chain-of-thought~\citep{wei2022chain} to enhance explainability. However, these text-based approaches often suffer from high inference latency and malformed outputs, whereas our ReCogDrive framework combines a VLM with a diffusion planner to produce more stable, safer trajectories while achieving a 7.8× speedup.
\noindent\textbf{Diffusion Models for Policy Learning.}
% Diffusion models have recently shown great potential in image generation~\cite{podell2023sdxl}, robotics~\cite{chi2023diffusion,liu2024rdt, black2024vision,yan2025m,bjorck2025gr00t}, and traffic simulation~\cite{yang2024diffusion,zhong2023guided}. Recently, many studies have introduced diffusion into autonomous driving. 
The recent success of diffusion models in domains like image generation~\citep{podell2023sdxl,zheng2024open}, robotics~\citep{chi2023diffusion,liu2024rdt, black2024vision,yan2025m,bjorck2025gr00t}, and traffic simulation~\citep{yang2024diffusion,zhong2023guided} has spurred their application in autonomous driving. For instance, DiffusionDrive~\citep{liao2024diffusiondrive} employs truncated diffusion with anchor priors for real-time multimodal planning. GoalFlow~\citep{xing2025goalflow} applies goal-point guidance with flow matching to generate high-quality trajectories. Diffusion Planner~\citep{zheng2025diffusion} redefines planning as a future trajectory generation task, jointly producing plans and motion forecasts. Our approach uses world knowledge from VLMs to guide the diffusion process, thereby improving the model's understanding of driving scenarios.

\noindent\textbf{Reinforcement Learning in Autonomous Driving.}
Reinforcement Learning has proven effective in games~\citep{hafner2023mastering}, robotics~\citep{ren2024diffusion}, and LLMs~\citep{guo2025deepseek}. Recently, many studies have introduced reinforcement learning into autonomous driving to improve model generalization. RAD~\citep{gao2025rad} proposes training an end-to-end AD agent using Reinforcement Learning in a photorealistic 3DGS environment. CarPlanner~\citep{zhang2025carplanner} learns an auto-regressive policy for consistent multi-modal trajectories, outperforming imitation learning. AlphaDrive~\citep{jiang2025alphadrive}, R2SE~\citep{liu2025reinforced} and TrajHF~\citep{li2025finetuning} incorporate GRPO to enhance the generalization of driving policies. Gen-Drive~\citep{huang2025gen} combines diffusion models with RL and reward modeling for better policies. We pioneer the application of reinforcement learning to Vision-Language-Action (VLA) models to enhance their planning capabilities.

\section{Preliminaries}

\paragraph{Problem Definition.}
Autonomous driving task aims to predict smooth and collision-free trajectory in future seconds, given the ego status $S_{\mathrm{ego}}$ (e.g., ego speed and ego acceleration), sensor input $I_{\mathrm{cam}}$, and navigation information $L_{\mathrm{nav}}$. Conventional end-to-end driving algorithm $\Phi$ formulate this as
\begin{equation}
    \label{eq:e2e_driving}
    \mathbf{V}_{\mathrm{traj}} = \Phi\bigl(I_{\mathrm{cam}},\,L_{\mathrm{nav}} ,\, S_{\mathrm{ego}}\bigr),
\end{equation}
where $\mathbf{V}_{\mathrm{traj}}\in\mathbb{R}^{T\times3}$ is the sequence of future waypoints and headings. While methods such as \citep{hu2023planning,jiang2023vad,liao2024diffusiondrive,chen2024vadv2} have shown strong effectiveness, their black-box nature impedes interpretability and they often fail to generalize to rare corner cases in real-world driving.

Recent works \citep{hwang2024emma,mao2023gpt,zhang2024instruct} utilize the rich world knowledge and causal reasoning capabilities of Vision–Language Models for autonomous driving. VLMs output trajectories in textual form and generate explicit reasoning processes:
\begin{equation}
    \label{eq:vlm_output}
    T_{\mathrm{traj}},\,T_{\mathrm{reason}} = \mathrm{VLM}\bigl(I_{\mathrm{cam}},\,L_{\mathrm{nav}} ,\, S_{\mathrm{ego}}\bigr).
\end{equation}
However, we observe an inherent mismatch between the language-formatted trajectory space and the continuous action space, and the autoregressive decoding process can suffer from output collapse, leading to erroneous trajectories.
% move below sentences to method section 
% The language space inherently suffers from precision limitations in numerical representation due to floating-point discretization constraints. 
% VLMs do not excel at precise numerical prediction.
% More importantly, VLMs exhibit sporadic hallucination artifacts, which undermine their reliability in safety-critical driving scenarios.
% Unlike VLMs, the Vision-Language-Action Model concatenates a VLM and an action decoder.
% \redtext{VLA methods, need to be repalced by citep}

% Inspired by~\citep{bjorck2025gr00t,wen2024diffusion,liu2025hybridvla,black2410pi0} in robotics, we decode a continuous trajectory from the hidden states in the last layer of VLM with a diffusion-based planner:
% % The hidden representation from the VLM and the ego-state are fed into the action decoder to generate the numeric trajectory:
% \begin{equation}
%     \label{eq:vla_reasoning}
%     T_{\mathrm{reason}},\,F_h = \mathrm{VLM}\bigl(I_{\mathrm{cam}},\,L_{\mathrm{nav}} ,\, S_{\mathrm{ego}}\bigr),
% \end{equation}
% \begin{equation}
%     \label{eq:vla_decoding}
%     \mathbf{V}_{\mathrm{traj}} = \Phi_{\mathrm{act}}\bigl(F_h,\,S_{\mathrm{ego}}\bigr),
% \end{equation}
% This paradigm retains the strong generalization and causal reasoning capabilities of the VLM while producing stable continuous trajectories.

\paragraph{Diffusion Policy.}

Denoising Diffusion Probabilistic Models (DDPMs)~\citep{ho2020denoising,nichol2021improved} learn a generative model by reversing a fixed, Markovian noising process that gradually corrupts data with Gaussian noise. Given a clean trajectory $\mathbf{x}_0$, the forward process defines:
\begin{equation}
    \label{eq:ddpm_train}
    q(\mathbf{x}_t \mid \mathbf{x}_{t-1})
    = \mathcal{N}\bigl(\mathbf{x}_t;\,\sqrt{1-\sigma_t^2}\,\mathbf{x}_{t-1},\,\sigma_t^2\mathbf{I}\bigr),\quad t=1,\dots,T,
\end{equation}
where $\sigma_t$ is the noise standard deviation at step $t$. At inference, trajectories are generated by initializing $\mathbf{x}_T \sim \mathcal{N}(\mathbf{0},\mathbf{I})$ and iteratively denoising:
\begin{equation}
\label{eq:ddpm_sample}
\mathbf{x}_{t-1}
= \frac{1}{\sqrt{1-\sigma_t^2}}\Bigl(\mathbf{x}_t - \sigma_t^2\,\epsilon_\theta(\mathbf{x}_t,t)\Bigr)
+ \sigma_t\,\mathbf{z},\quad
\mathbf{z}\sim\mathcal{N}(0,\mathbf{I}).
\end{equation}
Denoting trajectory waypoint as $\mathbf{x}_0$, this framework naturally extends to trajectory‐level policy generation, where the denoising network $\epsilon_\theta$ learns to refine noisy motion plans into smooth trajectories.

\section{Methodology}
% In this section, we present the three‐stage training paradigm of \name{}. First, we construct a 3M high‐quality driving QA dataset to pretrain a Vision–Language Model tailored for driving scenarios, endowing it with perception, prediction, and planning capabilities. Next, we integrate the pretrained VLM with a diffusion‐based planning decoder to achieve stable semantic‐to‐action mapping, translating latent language representations into continuous trajectories. Finally, we employ Expert‐Scorer–Guided Reinforcement Learning to fine‐tune the diffusion policy, aligning generated trajectories with human expert preferences.
% In this section, we present the overall framework and training paradigm of ReCogDrive. First, we assemble a driving dataset of 3.1 million high-quality QA pairs to pre-train Vision–Language Models, injecting driving-specific cognition into the model. Next, we integrate the pre-trained VLMs with a diffusion‐based trajectory planner to achieve stable semantic‐to‐action mapping, translating latent language space into continuous action space. Finally, we introduce simulation-assisted reinforcement learning to integrate generalized driving cognition, acquired through multi-trajectory exploration, into the diffusion planner.

In this section, we present the overall framework and training paradigm of ReCogDrive. First, we introduce a scalable hierarchical data pipeline that curates a multi-level cognitive dataset spanning from perception to reasoning to instill the VLM with nuanced, human-like driving priors. Second, we detail our cognition-guided diffusion planner, a novel architecture that resolves the modality mismatch by translating the VLM's latent semantic representations into continuous and stable trajectories. Finally, we describe our policy refinement stage, where a Diffusion Group Relative Policy Optimization (DiffGRPO) algorithm is employed to fine-tune the planner with simulated rewards, optimizing for safety and comfort beyond the confines of the expert dataset.

\begin{figure}[t!]
    \centering
    \includegraphics[width=\linewidth]{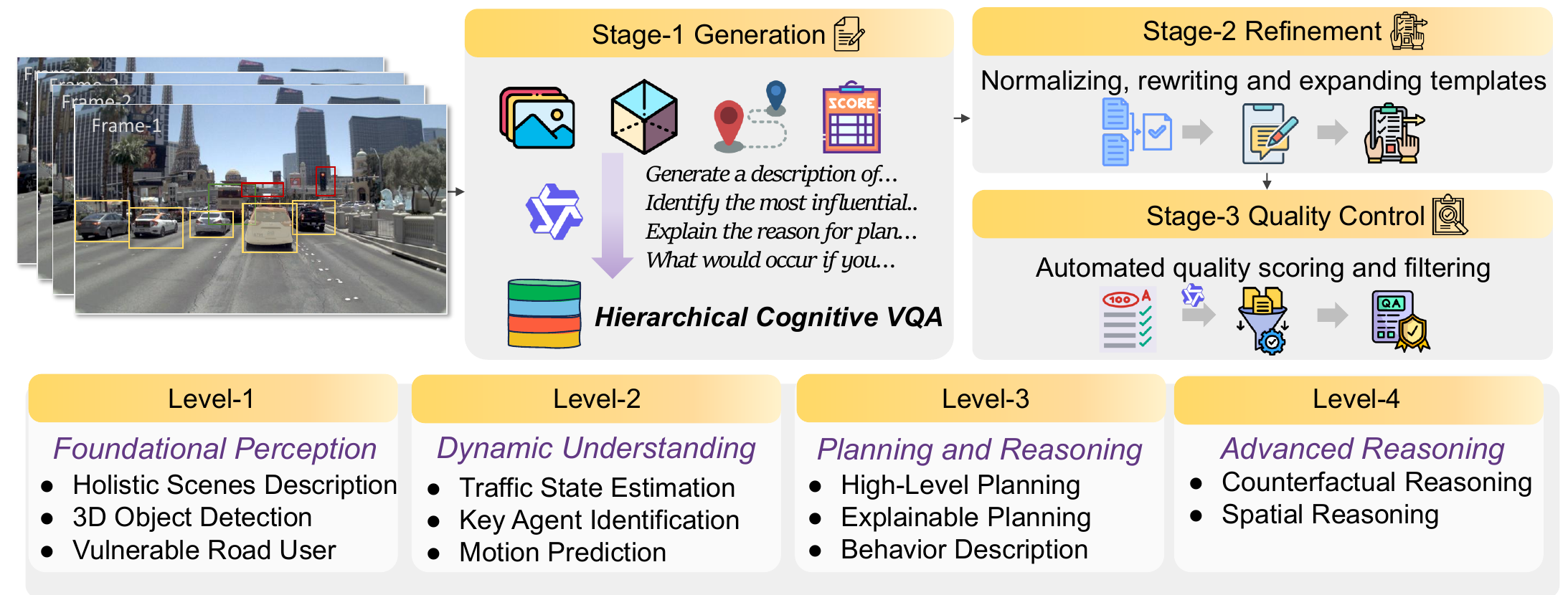}
    \caption{\textbf{Overview of our Scalable Hierarchical Data Pipeline.} We employ a three-stage process of Generation, Refinement, and Quality Control to produce a high-quality dataset that mimics the cognitive process of a human driver.}
    \label{fig:representation_gen}
    \vspace{-0.3cm}
\end{figure}

\subsection{Scalable Hierarchical Data Pipeline For Driving Pretraining}
To bridge the gap between general Vision-Language Models (VLMs) and autonomous driving, we introduce a scalable, structured pipeline for the automated generation of high-fidelity driving data, as shown in Fig.~\ref{fig:representation_gen}. This pipeline comprises three stages: Generation, Refinement, and Quality Control.

In the Generation stage, we align with the sequential cognitive process of human drivers, organized into four levels: \textit{(1) Foundational Perception:} constructing representations of static and dynamic scene elements; \textit{(2) Dynamic Understanding:} analyzing multi-agent dynamics and predicting near-term behaviors of surrounding entities; \textit{(3) Planning and Reasoning:} formulating executable, safety-aware driving plans and providing concise rationales; and \textit{(4) Advanced Reasoning:} conducting counterfactual analyses and fine-grained trade-off evaluations. To achieve this, we combine ground-truth data for objective tasks with a state-of-the-art VLM for subjective annotations.

The Refinement stage integrates open-source datasets, followed by normalization, rewriting, and question-template augmentation to ensure linguistic and semantic consistency. Finally, the Quality Control stage applies automated scoring and rigorous filtering based on linguistic accuracy, visual clarity, etc., retaining only the high quality samples. This systematic design ensures our pre-training dataset is both reliable and scalable, enabling the adaptation of a general VLM into a domain-specialized cognitive model for autonomous driving.

% First, the \textit{Generation} stage aligns with the sequential cognitive process of human drivers and is organized into the following four levels:
% \begin{enumerate}
%     \item \textbf{Foundational Perception:} Constructing representations of static and dynamic scene elements.
%     \item \textbf{Dynamic Understanding:} Interpreting complex interactions and predicting near-term behaviors of surrounding entities.
%     \item \textbf{Planning and Reasoning:} Formulating executable, safety-aware driving plans and providing concise rationales.
%     \item \textbf{Advanced Reasoning:} Conducting counterfactual analyses and fine-grained trade-off evaluations.
% \end{enumerate}

% Next, the Refinement stage enriches this corpus by merging it with open-source datasets, followed by normalization and rewriting to ensure consistency. Finally, the Quality Control stage applies automated scoring and rigorous filtering based on linguistic accuracy, visual clarity, etc., retaining only the high quality samples. This systematic design ensures our pre-training dataset is both reliable and scalable, enabling the adaptation of a general VLM into a domain-specialized cognitive model for autonomous driving.

\begin{figure}[t!]
    \centering
    \includegraphics[width=\linewidth]{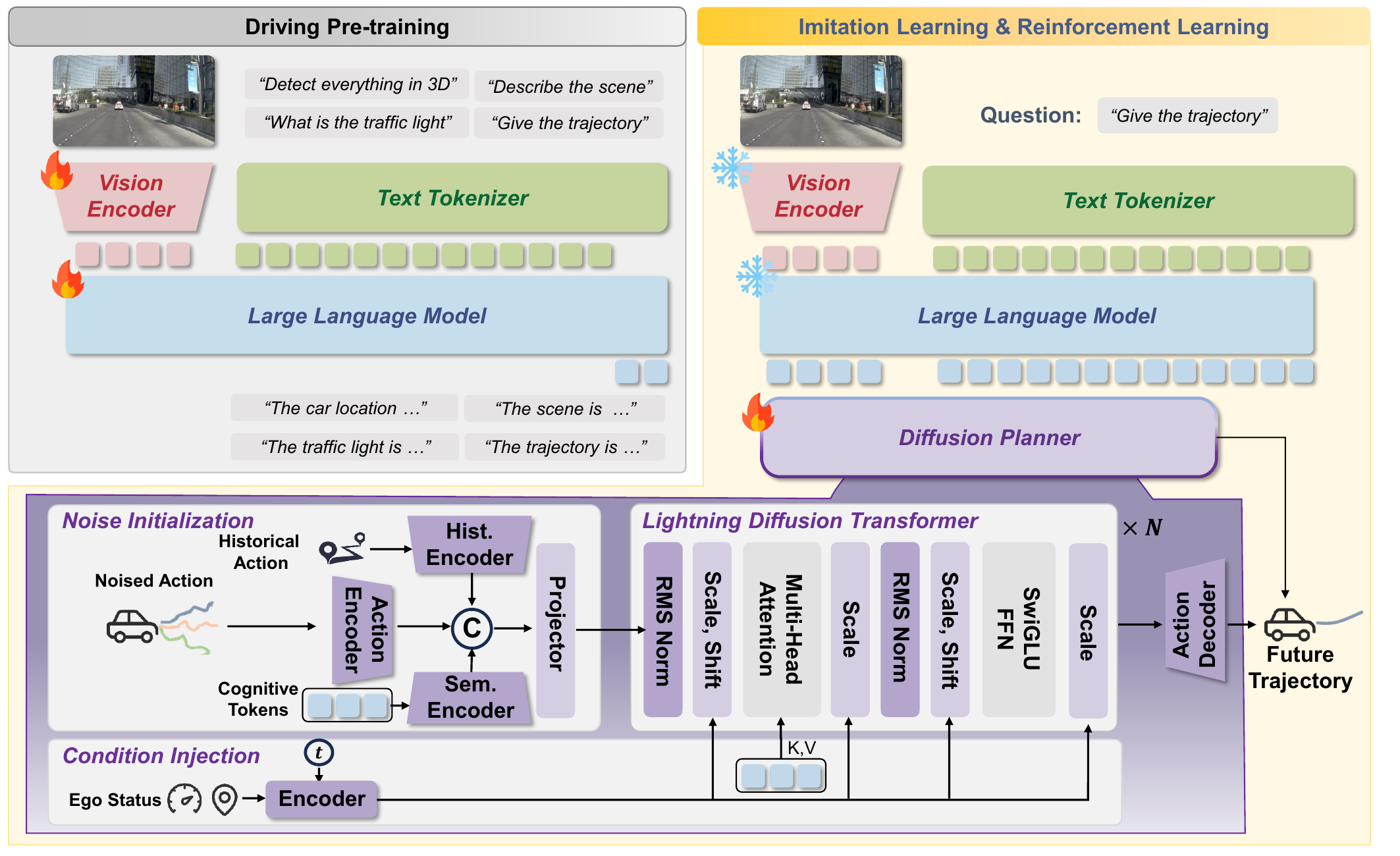}
    \caption{\textbf{Training Pipeline and Model Architecture.} ReCogDrive couples a VLM with a diffusion planner: the VLM encodes inputs into cognitive tokens guiding trajectory denoising. Training follows three stages: Driving Pre-training, imitation learning, and reinforcement learning.
}
    \label{fig:model_arch}
    \vspace{-0.3cm}
\end{figure}

\subsection{Cognition-Guided Diffusion Planner}

While autoregressive paradigms offer a straightforward way to generate trajectories in textual form, they suffer from inherent limitations: (i) the mismatch between discrete language space and continuous action space often leads to infeasible or jerky trajectories, and (ii) VLMs are prone to hallucination and decoding errors, which compromise safety in autonomous driving. To address these issues, we propose a \textit{Cognition-Guided Diffusion Planner}, which integrates human cognitive driving priors from a VLM with a diffusion planner to generate smooth and safety-aware trajectories. Formally, given a noisy trajectory sample $\mathbf{x}_t \in \mathbb{R}^{N \times 3}$, the denoising step is defined as:
\begin{equation}
\mathbf{x}_{t-1} = 
D_{\mathrm{act}}\Bigl(\mathrm{DiT}_\theta(z_t;\, F_h;\, S_{ego};\, t)\Bigr),
\quad 
z_t = \mathrm{concat}\!\left(E_{\mathrm{act}}(\mathbf{x}_t),\,
E_{\mathrm{his}}(\mathbf{x}_{\mathrm{hist}}),\,
\bar{F}_h \right),
\end{equation}

where $F_h$ denotes the VLM hidden states, $S_{ego}$ the ego-vehicle status, and $z_t$ the fusion of noisy actions, historical trajectories, and semantic priors. The diffusion model is trained by minimizing:
\begin{equation}
\mathcal{L}_{\text{dif}} = \mathbb{E}_{z_{t},c}\left\| \epsilon - \epsilon_{\pi}(z_{t}, c) \right\|^{2},
\end{equation}
with $\epsilon \sim \mathcal{N}(0, \mathbf{I})$ and condition $c=\{F_h, S_{ego}\}$.

% \textbf{Vision-Language Model Guidance} We adopt InternVL3~\citep{zhu2025internvl3,chen2024internvl} as the cognitive backbone for its strong multi-modal reasoning ability. Input images are partitioned into $448 \times 448$ patches with an additional thumbnail, encoded via InternViT, and compressed into 256 visual tokens through pixel-shuffle and projection layers. These tokens are concatenated with text tokens and processed by the language head, yielding final-layer hidden states $F_h \in \mathbb{R}^{L \times D}$, which are used as cross-attention conditions in the diffusion transformer. In addition, we apply average pooling to obtain a global semantic embedding $\bar{F}_h$, which is concatenated with the fused representation $z_t$ to provide stable contextual guidance during denoising. Moreover, the VLM can optionally output high-level driving instructions and chain-of-thought reasoning to further guide the diffusion process.

\textbf{Vision-Language Model Guidance.} We adopt InternVL3~\citep{zhu2025internvl3,chen2024internvl} as the cognitive backbone for its strong multi-modal reasoning. The VLM encodes multi-view images and textual queries into hidden states $F_h \in \mathbb{R}^{L \times D}$, which act as cognitive tokens carrying driving priors and serve as cross-attention conditions in the diffusion transformer. We further obtain a global semantic embedding $\bar{F}_h$ via average pooling over $F_h$, providing stable contextual guidance during denoising. Additionally, the VLM can optionally output high-level driving instructions and chain-of-thought reasoning to enhance trajectory generation.

\textbf{Diffusion Transformer Design.} Our diffusion planner builds on DiT~\citep{peebles2023scalable,yao2025vavae} blocks, enhanced with lightweight but effective design choices: SwiGLU FFN~\citep{shazeer2020glu} for expressive non-linearity, RoPE~\citep{su2024roformer} for relative positional encoding, and both QK-Norm and RMSNorm~\citep{zhang2019root} for stable optimization. Each block alternates between (i) \textit{self-attention}, which models pairwise waypoint relations, and (ii) \textit{cross-attention}, which injects semantic priors $F_h$ into the trajectory space. This design enables a structured fusion of scene-level cognition with low-level trajectory optimization.

\textbf{Historical and Ego Information Fusion.} To capture temporal consistency, historical trajectories $\mathbf{x}_{\mathrm{hist}}$ are encoded and concatenated with the noisy trajectory embeddings. Ego-vehicle status $S_{ego}$ (e.g., speed, acceleration) is injected via \textit{AdaLN modulation}, ensuring that the planner is conditioned on the vehicle’s physical state. This dual conditioning mechanism enhances stability, safety, and adaptability in diverse driving scenarios.

\begin{figure}[t!]
    \centering
    \includegraphics[width=\linewidth]{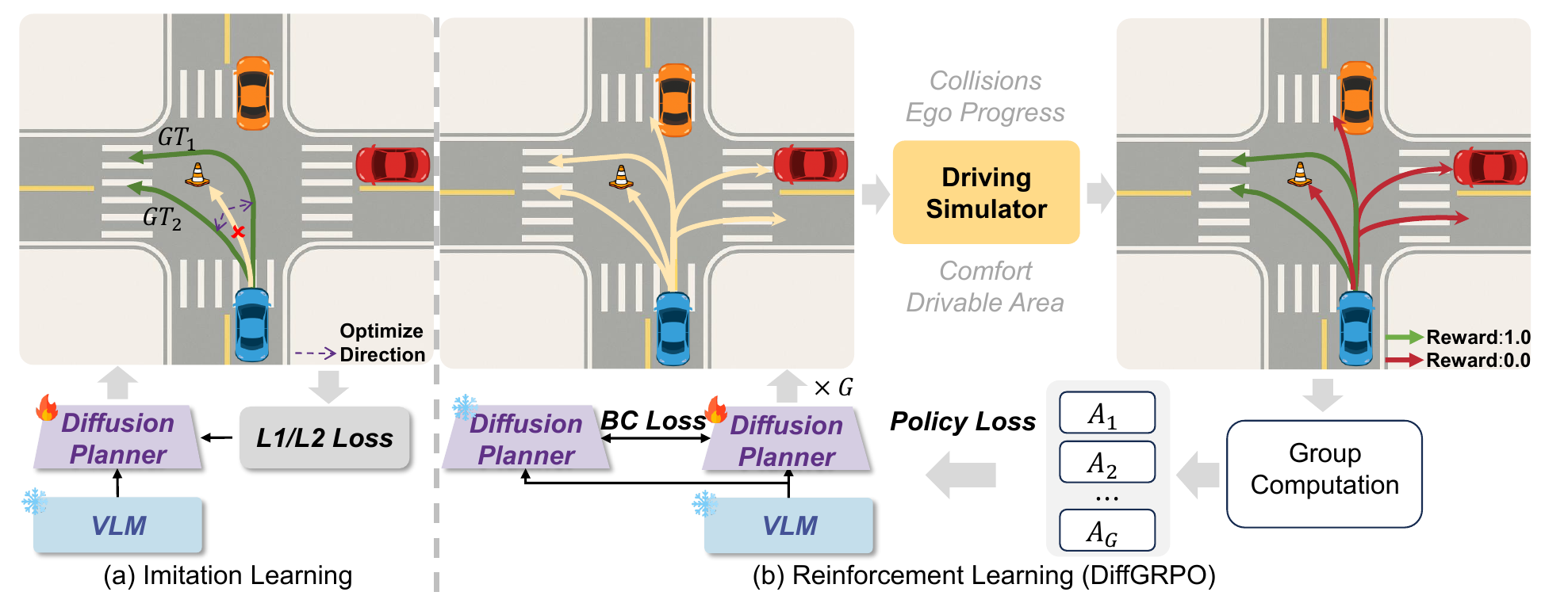}
    \caption{\textbf{Comparison of Training Paradigms.} (a) \textbf{Imitation Learning:} the diffusion planner is trained offline to mimic ground truth trajectories using L1/L2 losses, but tends to learn averaged, suboptimal paths. (b) \textbf{Reinforcement Learning:} multiple trajectories are sampled and evaluated in the NAVSIM simulator, scored on collision avoidance, drivable area compliance and other metrics, and advantages are computed via group computation to update the diffusion planner.}
    \label{fig:training_paradigms}
    \vspace{-0.3cm}
\end{figure}

\subsection{Diffusion Group Relative Policy Optimization}

Relying solely on imitation learning exhibits fundamental limitations~\citep{chu2025sft,osa2018algorithmic,ren2024diffusion}, since expert demonstrations may be multi-modal, leading to conflictual optimization results. 
As shown in Fig.~\ref{fig:training_paradigms}(a), when trained in this rare intersection turn scenario with multiple expert trajectories, the model resorts to learning an average trajectory to achieve global optimality, which can lead to incorrect or unsafe maneuvers. Consequently, even though the diffusion planner trained through imitation learning closely matches expert trajectories in terms of displacement, it still frequently produces collisions or drives outside the drivable area. 

To enable exploration beyond imitation, we introduce \textit{DiffGRPO}, a Group Relative Policy Optimization algorithm specifically designed for diffusion planners that tightly couples GRPO with the denoising process. 
% Unlike prior approaches~\citep{li2025finetuning} that employ simple $\ell_2$ trajectory distance as the reward, we are the first to integrate the NAVSIM simulator into reinforcement learning. The simulator evaluates each rollout in terms of collisions, comfort, and drivable-area compliance, and returns a Predictive Driver Model Score (PDMS) as the reward signal, as shown in Fig.~\ref{fig:training_paradigms}(b). This design grounds learning in realistic driving objectives rather than proxy losses. 
Following \cite{ren2024diffusion}, we interpret the diffusion policy $\pi_\theta$ as an internal Markov decision process: starting from Gaussian noise, the model gradually denoises to generate a full trajectory. Concretely, we sample \(G\) trajectories, each represented by a diffusion chain:
\begin{equation}
\mathbf{x} = (x_T,\,x_{T-1},\,\dots,\,x_0),
\end{equation}
where \(T\) is the total number of denoising steps and transitions follow:
\begin{equation}
x_T \sim \mathcal{N}(0, \mathbf{I}), 
\qquad
x_{t-1} \sim \pi_\theta(x_{t-1}\mid x_t).
\end{equation}

% However, constructing a fully interactive closed-loop simulator is highly challenging, so we use the simulator of NAVSIM to evaluate collisions, comfort, and other metrics for reinforcement learning, as shown in Fig.~\ref{fig:training_paradigms}(b).
% % Following ~\citep{ren2024diffusion,black2023training}, we employ reinforcement learning to fine-tune the diffusion policy, enhancing exploration in unfamiliar states and generalization to novel scenarios.
Unlike prior approaches~\citep{li2025finetuning,li2025drive} that rely on simple $\ell_2$ trajectory distances as surrogate rewards, we are the first, to our knowledge, to leverage the NAVSIM simulator to providing realistic feedback on safety and comfort. Each rollout is evaluated in terms of collisions, drivable-area compliance, and driving comfort, with the results aggregated into a Predictive Driver Model Score (PDMS) that serves as the reward \(r_i\). We cast this as a single-step decision, treating the whole trajectory as one composite action with its PDMS score as the reward. We then compute the group‐standardized advantage:
\begin{equation}
\hat A_{i} \;=\; \frac{r_i - \mathrm{mean}\bigl(r_{1..G}\bigr)}%
{\sqrt{\mathrm{var}\bigl(r_{1..G}\bigr)}},
\quad
i=1,\dots,G.
\end{equation}

Each conditional step in the diffusion chain is a Gaussian policy:
\begin{equation}
\pi_\theta\bigl(x_{t-1}\mid x_t\bigr)
= \mathcal{N}\!\Bigl(x_{t-1};\,\mu_\theta(x_t,t),\,\sigma_t^2 I\Bigr),
\end{equation}
where \(\mu_\theta(x_t,t)\) is the model‐predicted mean and \(\sigma_t^2 I\) the (fixed) covariance. Thus, the probability density of the full chain \(\mathbf{x}_{0:T}\) under \(\pi_\theta\), where \(p(x_T)\) is a Gaussian prior independent of \(\theta\), is:
\begin{equation}
\log \pi_\theta(\mathbf{x}_{0:T})
= \log p(x_T) + \sum_{t=1}^T \log \pi_\theta(x_{t-1}\mid x_t),
\end{equation}

Finally, we compute the policy loss following~\citep{williams1992simple,shao2024deepseekmath} while concurrently incorporating a behavior cloning loss to prevent collapse during exploration.
\begin{equation}
\label{eq:grpo}
L \;=\;
\underbrace{-\frac{1}{G}\sum_{i=1}^G\frac{1}{T}\sum_{t=1}^T
\gamma^{\,t-1}\,\log \pi_\theta\bigl(x_{t-1}^{(i)}\mid x_t^{(i)}\bigr)\,\hat A_{i}}_{L_{\mathrm{RL}}}
\;-\;
\underbrace{\lambda\;\frac{1}{G}\sum_{i=1}^G\frac{1}{T}\sum_{t=1}^T
\log \pi_\theta\bigl(\tilde x_{t-1}^{(i)}\mid \tilde x_t^{(i)}\bigr)}_{ L_{\mathrm{BC}}},
\end{equation}
where $\gamma$ is the discount coefficient mitigating instability in early denoising steps, $\lambda$ is the weight for the behavior cloning loss, and $\tilde x_{t-1}, \tilde x_t$ are values sampled from the reference policy $\pi_{\mathrm{ref}}$. We omit the PPO-style clipping and set the update iteration to 1, so $\pi = \pi_{\mathrm{old}}$.
% Note that we set the update iteration to 1 so that $\pi$ = $\pi_{\mathrm{old}}$ and do not use the PPO or the original GRPO clipping mechanism.
Through DiffGRPO, the diffusion planner learns to generate safe and comfortable trajectories in closed-loop settings, going beyond mere imitation and enhancing the robustness of our framework.

\section{Experiments} 

\label{expr}

\subsection{Experimental Setup.}  
\paragraph{Implementation Details.}
We choose InternVL3~\citep{zhu2025internvl3}, comprising a 300M-parameter InternViT visual encoder~\citep{chen2024internvl} and a Qwen2.5 Large-Language-Model~\citep{Qwen2.5-VL}, as our base model, which demonstrates strong performance on multiple benchmarks. Images are processed through a dynamic resolution preprocessing strategy. In the first stage, we conduct supervised fine-tuning (SFT) on the VLM using the dataset constructed by our hierarchical data pipeline for three epochs. In the second stage, with the VLM parameters frozen, we train the diffusion model via DDPM for 200 epochs. In the third stage, we further optimize the diffusion model using reinforcement learning for 10 epochs, with one policy update per batch. Detailed hyperparameter settings are provided in the supplementary material.

\paragraph{Dataset.}
We evaluate primarily on two challenging benchmarks: NAVSIM~\citep{dauner2025navsim} and Bench2Drive~\citep{jia2024bench2drive}.
NAVSIM is a planning-oriented autonomous driving dataset built on OpenScene~\citep{openscene2023}, a redistribution of nuPlan~\citep{caesar2021nuplan}.  The dataset is split into \textit{navtrain} (1,192 training scenes) and \textit{navtest} (136 evaluation scenes). Bench2Drive is a CARLA-based benchmark composed of 220 short routes, each containing a distinct, safety-critical scenario. In addition, to build the VLM's cognitive foundation, ReCogDrive is trained on a collection of VQA datasets, including DriveLM~\citep{sima2024drivelm} and LingoQA~\citep{marcu2024lingoqa}, with further details provided in the supplementary material.

\begin{table*}[t!]
    \centering
    \small
    \caption{\textbf{Performance comparison on NAVSIM \textit{navtest} using closed-loop metrics.} $^{\dagger}$ denotes models fine-tuned on the NAVSIM trajectory dataset.}
    \setlength{\tabcolsep}{4pt}
    \begin{tabular}{@{}l|cc|cc|ccc|cc@{}}
        \toprule
        Method &  Image & Lidar & NC$\uparrow$ & DAC$\uparrow$ & TTC$\uparrow$ & Comf. $\uparrow$ & EP$\uparrow$ & PDMS$\uparrow$ \\
        \midrule
        Constant Velocity & &   & 68.0 & 57.8 & 50.0 & \textbf{100} & 19.4 & \cellcolor{gray!30} 20.6 \\
        Ego Status MLP &  &  & 93.0 & 77.3 & 83.6 & \textbf{100} & 62.8 & \cellcolor{gray!30} 65.6 \\        
        \midrule
        VADv2-$\mathcal{V}_{\text{8192}}$~\citep{chen2024vadv2} & \checkmark &  &  97.2 & 89.1 & 91.6 & \textbf{100} & 76.0 &\cellcolor{gray!30}  80.9 \\
        DrivingGPT~\citep{chen2024drivinggpt} & \checkmark &  &  \textbf{98.9} & 90.7 & 94.9 & 95.6 & 79.7 & \cellcolor{gray!30} 82.4 \\ 
        % Hydra-MDP-$\mathcal{V}_{\text{8192}}$ & \checkmark & \checkmark &   97.9 & 91.7 & 92.9 & \textbf{100} & 77.6 & \cellcolor{gray!30} 83.0 \\
        UniAD~\citep{hu2023planning} & \checkmark &    & 97.8 & 91.9 & 92.9 & \textbf{100} & 78.8 & \cellcolor{gray!30} 83.4 \\
        % LTF~\citep{chitta2022transfuser} & \checkmark &    & 97.4 & 92.8 & 92.4 & \textbf{100} & 79.0 & \cellcolor{gray!30} 83.8 \\
        % BevDrive~\citep{yu2025combining} & \checkmark & \checkmark   & 97.7 & 92.5 & 92.9 & \textbf{100} & 78.7 & \cellcolor{gray!30} 83.8 \\
        TransFuser~\citep{chitta2022transfuser} & \checkmark & \checkmark   & 97.7 & 92.8 & 92.8 & \textbf{100} & 79.2 & \cellcolor{gray!30} 84.0 \\
        PARA-Drive~\citep{weng2024drive} & \checkmark &  & 97.9 & 92.4 & 93.0 & 99.8 & 79.3 & \cellcolor{gray!30} 84.0 \\
        DRAMA ~\citep{yuan2024drama} & \checkmark & \checkmark   & 98.0 & 93.1 & 94.8 & 100 & 80.1 & \cellcolor{gray!30} 85.5 \\
        Hydra-MDP-$\mathcal{V}_{\text{8192}}$-W-EP~\citep{li2024hydra} & \checkmark & \checkmark  & 98.3 & 96.0 & 94.6 & \textbf{100} & 78.7 & \cellcolor{gray!30} 86.5 \\
        % ARTEMIS~\citep{feng2025artemis} & \checkmark & \checkmark & 98.3 & 95.1 & 94.3 & \textbf{100} & 81.4 & \cellcolor{gray!30} 87.0 \\
        DiffusionDrive~\citep{liao2024diffusiondrive} & \checkmark & \checkmark & 98.2 & 96.2 & 94.7 & \textbf{100} & 82.2 & \cellcolor{gray!30} 88.1 \\
        WoTE~\citep{li2025end} & \checkmark & \checkmark & 98.5 & 96.8 & \textbf{94.9} & 99.9 & 81.9 & \cellcolor{gray!30} 88.3 \\
        \midrule
        \multicolumn{8}{@{}l}{\raggedright \textbf{VLMs-based Methods}} \\
        QwenVL2.5-8B$^{\dagger}$~\citep{Qwen2.5-VL} & \checkmark &  &  97.8 & 92.1 & 92.8 & \textbf{100} & 78.3 & \cellcolor{gray!30} 83.3 \\
        InternVL3-8B$^{\dagger}$~\citep{zhu2025internvl3} & \checkmark &  &  97.0 & 92.4 & 91.8 & \textbf{100} & 78.9 & \cellcolor{gray!30} 83.3 \\
        \textit{ReCogDrive(ours)} & \checkmark &  &  97.9 & \textbf{97.3} & \textbf{94.9} & \textbf{100} & \textbf{87.3} & \cellcolor{gray!30} \textbf{90.8} \\
        % \textit{ReCogDrive-8B(ours)} & \checkmark &  &  97.8 & \textbf{97.7} & \textbf{94.9} & \textbf{100} & \textbf{86.3} & \cellcolor{gray!30} \textbf{90.5} \\
        \bottomrule
    \end{tabular}
    \label{tab:comparison_modified}
\end{table*}
% It features a comprehensive sensor suite, including eight cameras for a 360° FOV and a merged LiDAR point cloud from five sensors. Annotations are provided at 2Hz, encompassing both HD maps and object bounding boxes. The dataset focuses on challenging driving scenarios involving dynamic changes in driving intentions, excluding trivial situations like stationary scenes or constant-speed driving. This emphasis on complex scenarios makes NAVSIM a valuable resource for developing and testing advanced autonomous driving algorithms.

\begin{table*}[t]
\centering
\caption{\textbf{Closed-loop and Multi-ability Testing Results in CARLA Bench2Drive Leaderboard.}}
\label{tab:bench2drive_full}
\small
\resizebox{\linewidth}{!}{
\begin{tabular}{l|ccc>{\columncolor[gray]{0.9}}c|ccccc>{\columncolor[gray]{0.9}}c}
\toprule
\multirow{2}{*}{\textbf{Method}} & \multicolumn{4}{c|}{\textbf{Closed-loop Metric $\uparrow$}} & \multicolumn{5}{c}{\textbf{Multi-Ability Test} (\%) $\uparrow$} \\
\cmidrule{2-11}
& Efficiency & Comfort& Success& \textbf{DS} & Merging & Overtaking & Emerg. Brake & GiveWay & Traf. Sign & \textbf{Mean} \\
\midrule
TCP*~\citep{wu2022trajectory} & 54.26 & \underline{47.80} & 15.00 & 40.70 & 16.18 & 20.00 & 20.00 & 10.00 & 6.99 & 14.63 \\
TCP-ctrl*~\citep{wu2022trajectory} & 55.97 & \textbf{51.51} & 7.27 & 30.47 & 10.29 & 4.44 & 10.00 & 10.00 & 6.45 & 8.23 \\
TCP-traj*~\citep{wu2022trajectory} & 76.54 & 18.08 & 30.00 & 59.90 & 8.89 & 24.29 & {51.67} & \underline{40.00} & 46.28 & 34.22 \\
TCP-traj w/o distill.~\citep{wu2022trajectory} & 78.78 & 22.96 & 30.05 & 49.30 & 17.14 & 6.67 & 40.00 & \underline{40.00} & 28.72 & 28.51 \\
ThinkTwice~\citep{jia2023think} & 76.93 & 16.22 & 3.13 & 62.44 & 27.38 & 18.42 & 35.82 & \textbf{50.00} & 54.23 & 37.17 \\
DriveAdapter*~\citep{jia2023driveadapter} & 70.22 & 16.01 & 33.08 & \underline{64.22} & \underline{28.82} & \underline{26.38} & \underline{48.76} & \textbf{50.00} & \underline{56.43} & \textbf{42.08} \\\midrule
AD-MLP~\citep{zhai2023rethinking} & 48.45 & 22.63 & 0.00 & 18.05 & 0.00 & 0.00 & 0.00 & 0.00 & 4.35 & 0.87 \\
UniAD-T.~\citep{hu2023planning} & 123.92 & 47.04 & 13.18 & 40.73 & 8.89 & 9.33 & 20.00 & 20.00 & 15.43 & 14.73 \\
UniAD-B.~\citep{hu2023planning} & 129.21 & 43.58 & 16.36 & 45.81 & 14.10 & 17.78 & 21.67 & 10.00 & 14.21 & 15.55 \\
VAD~\citep{jiang2023vad} & \textbf{157.94} & 46.01 & 15.00 & 42.35 & 8.11 & 24.44 & 18.64 & 20.00 & 19.15 & 18.07 \\
DriveTransformer-L.~\citep{jia2025drivetransformer} & 100.64 & 46.01 & \underline{35.01} & 63.46 & 17.57 & \textbf{35.00} & 48.36 & \underline{40.00} & 52.10 & 38.60 \\
\midrule
\textbf{ReCogDrive} & \underline{138.18} & 17.45 & \textbf{45.45} & \textbf{71.36} & \textbf{29.73} & 20.00 & \textbf{69.09} & 20.00 & \textbf{71.34} & \underline{42.03} \\
\bottomrule
\end{tabular}}
\vspace{-0.3cm}
\label{tab:carla_b2d}
\end{table*}

\subsection{Main Results and Ablation Study}

\paragraph{Experiments on the NAVSIM Benchmark.}
Tab.~\ref{tab:comparison_modified} reports results on NAVSIM. \name{} achieves a PDMS of 90.8, establishing a new state-of-the-art. Remarkably, it outperforms DiffusionDrive~\citep{liao2024diffusiondrive} and WoTE~\citep{li2025end}, both of which use camera and LiDAR inputs, by 2.7 and 2.5 PDMS, respectively, while using only camera input. Compared with our reproduced InternVL3~\citep{zhu2025internvl3} and QwenVL2.5~\citep{Qwen2.5-VL} trained directly on NAVSIM trajectories, \name{} improves PDMS by 7.5, validating the effectiveness of our training paradigm. It also surpasses the prior vision-only state-of-the-art, PARA-Drive~\citep{weng2024drive}, by 6.8 PDMS.

\paragraph{Bench2Drive Closed-loop Performance.} 

\cref{tab:bench2drive_full} reports closed-loop and multi-ability results on the CARLA Bench2Drive leaderboard. \name{} achieves the highest scenario success rate of 45.45\% and the top Driving Score of 71.36, surpassing prior end-to-end baselines. It also excels in safety-critical skills such as emergency braking 69.09\% and traffic sign compliance 71.34, while maintaining strong efficiency and a competitive multi-ability mean of 42.03. These results highlight the effectiveness and reliability of our framework in complex urban driving.

% \begin{wraptable}[9]{r}{0.46\textwidth} 
% \vspace{-0.03cm}
% \centering
% \caption{Model performance on DriveLM and DriveBench.}
% \label{tab:drive_results}
% \vspace{-0.3cm}
% \scriptsize
% \setlength{\tabcolsep}{1.5pt} % 缩小列间距
% \begin{tabular}{l | c | c c c c | c}
% \toprule[0.8pt]
% \multirow{2}{*}{Method} & DriveLM & \multicolumn{5}{c}{DriveBench} \\
% \cmidrule(lr){2-2} \cmidrule(lr){3-7}
%  & GPT-Score & Percep. & Predict. & Plan. & Behav. & Avg. \\
% \toprule[0.4pt]
% LLaVA-1.5   & 61.91 & 23.22 & 22.02 & 29.15 & 13.60 & 22.00 \\
% InternVL2   & 64.13 & 32.36 & 45.52 & 53.27 & 54.58 & 46.43 \\
% Qwen2-VL    & -- & 30.13 & 49.35 & 61.30 & 51.26 & 48.01 \\
% DriveLM     & 65.25 & 16.85 & 44.33 & 68.71 & 42.78 & 43.17 \\
% Dolphins    & -- & 9.59 & 32.66 & 52.91 & 8.81 & 25.99 \\
% GPT-4o      & 67.27 & 35.37 & 51.30 & 75.75 & 45.40 & 51.96 \\
% \midrule
% ReCogDrive  & \textbf{67.30} & 64.95 & 49.34 & 70.20 & 42.36 & \textbf{56.71} \\
% \bottomrule
% \end{tabular}
% \end{wraptable}

\begin{table}[t]
  \centering
  \small
\caption{\textbf{Ablation study on the proposed components of \name{}.} We evaluate the effect of driving pre-training, diffusion planner, and reinforcement learning on NAVSIM evaluation.}
  \setlength{\tabcolsep}{3pt}
  \vspace{-0.3cm}
  \renewcommand{\arraystretch}{0.9}
  \scalebox{1.1}{ 
  \begin{tabular}{
    c
    c c c c
    c c c c c | c
  }
    \toprule
    ID 
      & \makecell{Trajectory\\training} 
      & \makecell{Driving\\Pre-training} 
      & \makecell{Diffusion\\Planner} 
      & \makecell{Diff\\GRPO} 
      & NC  
      & DAC 
      & TTC 
      & Conf. 
      & EP  
      & \cellcolor{gray!30} PDMS \\
    \midrule
    1 
      & \ding{51} & \ding{55} & \ding{55} & \ding{55}
      & 97.4 & 91.3 & 93.0 & 100 & 77.2 & \cellcolor{gray!30} 82.4 \\
    2 
      & \ding{51} & \ding{51} & \ding{55} & \ding{55}
      & 97.6 & 93.1 & 92.7 & 100 & 79.1 & \cellcolor{gray!30} 84.1 \\
    % 3 
    %   & \ding{55} & \ding{55} & \ding{51} & \ding{55}
    %   & 96.8 & 91.3 & 90.2 & 100 & 65.4 & \cellcolor{gray!30} 76.3 \\
    3 
      & \ding{51} & \ding{51} & \ding{51} & \ding{55}
      & 98.1 & 94.7 & 94.2 & 100 & 80.9 & \cellcolor{gray!30} 86.5\\
    4 
      & \ding{51} & \ding{51} & \ding{51} & \ding{51}
      & 97.9 & 97.3 & 94.9 & 100.0 & 87.3 & \cellcolor{gray!30} 90.8 \\
    \bottomrule
  \end{tabular}}
  \vspace{3pt}
  \vspace{-15pt}
  \label{tab:abl}
\end{table}

\begin{table*}[t]
  \centering
  \begin{minipage}[t]{0.48\linewidth}
    \centering
    \caption{Comparison of trajectory generation methods.$\dagger$ indicates inference with \texttt{lmdeploy}.}
    \label{tab:vlm_trajectory}
    \vspace{-0.3cm}
    \scriptsize
    \setlength{\tabcolsep}{4pt} % 调整列间距
    \renewcommand{\arraystretch}{1.13}
    \begin{tabular}{l | c c >{\columncolor{red!10}}c}
    \toprule[0.8pt]
    Method & Time (s) & PDMS $\uparrow$ & Err. (\%) $\downarrow$ \\
    \toprule[0.4pt]
    VLM Plain Text$^{\dagger}$         & 0.5839 & 84.1 & 0.01 \\
    \midrule
    VLM + MLP                    & 0.0751 & 75.5 & 0.00 \\
    VLM + Query-based Decoder        & 0.0750 & 85.0 & 0.00 \\
    \midrule
    VLM + Vanilla Diffusion    & 0.0748 & 85.4 & 0.00 \\
    \quad \textit{+ SwiGLU FFN}    & 0.0745 & 85.7 & 0.00 \\
    \quad \textit{+ RMS Norm}    & 0.0749 & 85.8 & 0.00 \\
    \quad \textit{+ RoPE \& QK Norm} & 0.0750 & 86.5 & 0.00\\
    \bottomrule
    \end{tabular}
  \end{minipage}%
  \hfill
  \begin{minipage}[t]{0.48\linewidth}
    \centering
    \caption{Model performance on Drive VQA Benchmark.}
    \label{tab:drive_results}
    \vspace{-0.3cm}
    \scriptsize
    \setlength{\tabcolsep}{1.5pt} % 缩小列间距
    \begin{tabular}{l | c | c c c c | c}
    \toprule[0.8pt]
    \multirow{2}{*}{Method} & DriveLM & \multicolumn{5}{c}{DriveBench} \\
    \cmidrule(lr){2-2} \cmidrule(lr){3-7}
     & GPT-Score & Percep. & Predict. & Plan. & Behav. & Avg. \\
    \toprule[0.4pt]
    LLaVA-1.5   & 61.91 & 23.22 & 22.02 & 29.15 & 13.60 & 22.00 \\
    InternVL2   & 64.13 & 32.36 & 45.52 & 53.27 & 54.58 & 46.43 \\
    Qwen2-VL    & -- & 30.13 & 49.35 & 61.30 & 51.26 & 48.01 \\
    DriveLM     & 65.25 & 16.85 & 44.33 & 68.71 & 42.78 & 43.17 \\
    Dolphins    & -- & 9.59 & 32.66 & 52.91 & 8.81 & 25.99 \\
    GPT-4o      & 67.27 & 35.37 & 51.30 & 75.75 & 45.40 & 51.96 \\
    \midrule
    ReCogDrive  & \textbf{67.30} & 64.95 & 49.34 & 70.20 & 42.36 & \textbf{56.71} \\
    \bottomrule
    \end{tabular}
    % \caption{Effect of QA dataset scale and quality on planning. “LQ” and “HQ” denote low-quality and high-quality data, respectively.}
    \label{tab:training-qa-growth}
  \end{minipage}%
\end{table*}

\begin{figure*}[t!]
    \centering
    \includegraphics[width=0.95\linewidth]{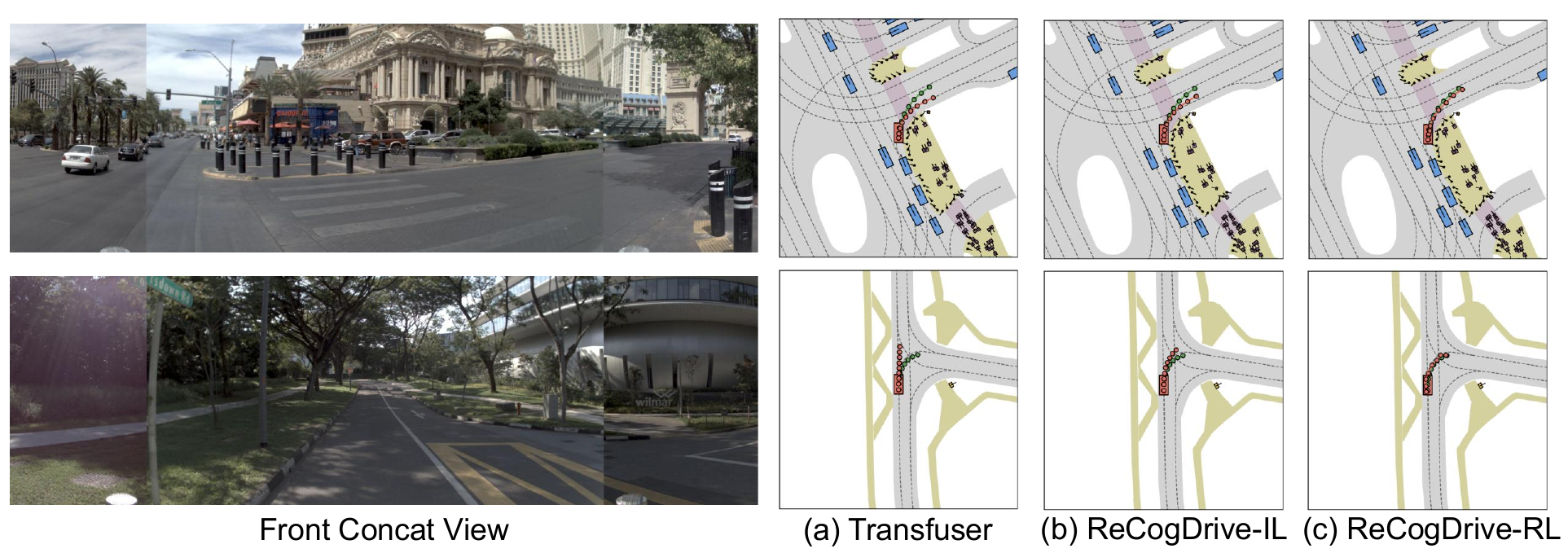}
    \vspace{-5pt}
    \caption{\textbf{Comparisons on the Navtest benchmark.}}
    \label{fig:qualitative}
    \vspace{-8pt}
\end{figure*}

\paragraph{Ablation study on \name{}.}
% As shown in Tab.~\ref{tab:abl}, we incrementally add each proposed component to evaluate its contribution. Starting with only trajectory training, we observe a notable improvement by incorporating driving pre-training, which highlights the importance of adapting VLMs to the driving domain. Introducing the diffusion decoder further enhances planning performance by enabling continuous trajectory generation. Finally, reinforcement learning brings the model to its best performance, achieving a PDMS of 89.6, demonstrating the effectiveness of Expert–Guided reinfocement learning in producing safer and more human-aligned driving behavior.
Tab.~\ref{tab:abl} presents an ablation study on the proposed components of \name{}. When training InternVL3 solely on NAVSIM trajectory data, the model achieves a PDMS of 83.3. Adapting the VLM to driving scenarios with our large-scale driving QA data increases PDMS by 1.7. Introducing the diffusion planner for continuous trajectory prediction further raises PDMS by 2.4. Finally, DiffGRPO achieves a PDMS of 90.8 with a 4.3 improvement, demonstrating the effectiveness of our RL scheme in producing safer driving behavior.

% \paragraph{Comparison of Trajectory Generation Methods.}
% We compare various trajectory generation methods in terms of inference speed, PDM Score, and error rate, with results presented in Table~\ref{tab:vlm_trajectory}. Despite its reasonable performance, generating trajectories as plain text is slow due to its autoregressive nature and poses critical safety risks from potential format errors. In contrast, methods using MLP, Query-based Decoder, and diffusion planner are substantially faster. Among these, diffusion-based approaches demonstrate a clear advantage in improving the PDM Score. To further enhance performance, we incrementally integrated architectural improvements including SwiGLU FFN, RMS Norm, RoPE, and QK Norm. Our final model, ReCogDrive, which incorporates these enhancements, achieves a 7.8x speedup and a 2.4-point increase in PDMS over the plain text baseline, validating the effectiveness of our proposed method.
\paragraph{Comparison of Trajectory Generation Methods.}
We compare trajectory generation methods in terms of inference speed, PDM Score, and error rate as shown in Tab.~\ref{tab:vlm_trajectory}. Generating trajectories as plain text is slow due to the autoregressive process and may suffer from format errors that compromise safety. In contrast, MLP, Query-based Decoder~\citep{chitta2022transfuser}, and diffusion planner are substantially faster, with diffusion-based approaches further boosting PDMS. Building on this, we integrated lightweight architectural refinements including SwiGLU FFN, RMS Norm, RoPE, and QK Norm. Our final model ReCogDrive achieves a 7.8$\times$ speedup and a 2.4 PDMS gain over the text baseline, validating the effectiveness of our design.

\paragraph{Evaluation on Drive VQA Benchmarks.}
To provide a comprehensive assessment of our model's language-related capabilities in driving scenarios, we evaluate its performance on the DriveLM and DriveBench benchmarks. The results, presented in Tab.~\ref{tab:drive_results}, demonstrate the effectiveness of our approach. Our model, ReCogDrive, achieves state-of-the-art performance, surpassing the powerful closed-source GPT-4o model on most key metrics, including the overall DriveBench average score. Furthermore, ReCogDrive significantly outperforms several specialized open-source driving models, validating its strong visual question answering and reasoning abilities in this domain.

\paragraph{Qualitative Results.} 
In Fig.~\ref{fig:qualitative}, we compare ReCogDrive (IL and RL) with Transfuser~\citep{chitta2022transfuser}, where RL yields safer and more reliable trajectories in challenging turning scenarios. More visualizations are in the supplementary material.

\section{Conclusion}

% In this work, we propose \name{}, an end-to-end autonomous driving system that integrates \ac{vlms} with diffusion-based trajectory planner, along with a three-stage training paradigm. 
% First, we assemble and curate a 3.1M high-quality driving QA dataset to inject driving-specific cognition into a pre-trained VLM.
% Second, we train a diffusion-based trajectory planner via DDPM to map discrete language representations into smooth, continuous trajectories, while preserving the VLM’s inherent world cognition and driving-specific cognition.
% Finally, we reinforce the diffusion policy to integrate generalized driving cognition into the diffusion planner.
% Extensive experiments on NAVSIM demonstrate that \name{} achieves state-of-the-art on closed-loop metrics without LiDAR input. 

% In this work, we proposed \name{}, a novel Reinforced Cognitive framework for end-to-end autonomous driving that integrates a Vision-Language Model with a diffusion planner. We first introduced a scalable hierarchical data pipeline that mimics human driving cognition to construct a foundation model with strong driving priors. Building on this, we designed a cognition-guided diffusion planner that leverages these priors for trajectory generation. Finally, we applied DiffGRPO to fine-tune the planner, enhancing safety and comfort. Extensive experiments on multiple benchmarks demonstrate that \name{} achieves state-of-the-art performance, validating the effectiveness of our approach.

In this work, we proposed \name{}, a novel reinforced cognitive framework for end-to-end autonomous driving that integrates a Vision-Language Model with a diffusion planner. We first introduced a scalable hierarchical data pipeline that mimics human driving cognition to construct a foundation model with strong driving priors. Building on this foundation, we designed a cognition-guided diffusion planner that injects VLM-derived cognitive tokens into the denoising process, enabling the generation of continuous, stable, and human-like trajectories. To further refine planning behavior, we introduced DiffGRPO, a reinforcement learning scheme that explicitly optimizes safety and comfort in closed-loop driving. Extensive experiments on multiple benchmarks demonstrate that \name{} achieves state-of-the-art performance, validating the effectiveness of our approach. We hope our work can advance the development of autonomous driving VLAs.

\bibliography{iclr2026_conference}
\bibliographystyle{iclr2026_conference}

\clearpage
\appendix
\section{Appendix}

\renewcommand{\thesection}{\Alph{section}}
We organize the supplementary material as follows. First, in \cref{sec:questions}, we address potential questions that may arise from reading the main text. We then report \name{}’s performance on the NAVSIM and DriveBench benchmarks, along with more detailed ablation studies in \cref{sup:results}. In \cref{sup:dataset_construction}, we provide details of the training data collection and processing pipeline. \cref{sup:implementation_details} describes \name{}’s training and inference implementation, including all key hyperparameters, and also provides details of the evaluation metrics used on NAVSIM and Bench2Drive. \cref{sup:limitations} and \cref{sup:broader_impacts} discuss the limitations of the method and the broader impacts. We also provide our LLM Usage Statement in \cref{sup:llm_usage}. Finally, \cref{sup:qualitative} presents additional qualitative results, including extensive visualizations on NAVSIM and Bench2Drive, comparisons between IL and RL trajectories, driving dialogues, and analyses of failure cases.

\section{Questions}
\label{sec:questions}

\noindent\textbf{Q1.} \textit{What is the technical novelty in this paper?}

The key novelty of this work is the first reinforced cognitive framework for autonomous driving that integrates a VLM with a diffusion-based planner. Unlike prior driving VLA approaches (e.g., EMMA~\citep{hwang2024emma}, OmniDrive~\citep{wang2024omnidrive}) that generate trajectories purely in the text space, our framework injects VLM-derived driving priors into a diffusion planner, avoiding slow inference, infeasible actions, and trajectory errors. Within this unified framework, we further contribute three innovations: (i) a scalable hierarchical data pipeline that mimics the sequential cognitive process of human drivers through generation, refinement, and quality control, enabling flexible adaptation of VLMs to diverse driving scenarios; (ii) a cognition-guided diffusion planner that leverages VLM-derived priors to efficiently generate continuous and stable trajectories, surpassing plain-text and Transfuser-like~\citep{chitta2022transfuser} decoders; and (iii) Diffusion Group Relative Policy Optimization (DiffGRPO), the first reinforcement learning scheme tailored for diffusion-based planners in autonomous driving VLAs, enabling simulation-driven policy optimization for safer and more comfortable driving.

\smallskip

\bigskip
\noindent\textbf{Q2.} \textit{Does the VLM merely act as a feature extractor?}

ReCogDrive consists of a Vision-Language Model and a diffusion planner. In our framework, the VLM can produce high-level instructions or chain-of-thought reasoning to guide planning. However, as shown in Tab.~\ref{tab:traj_com}, reinforcement learning alone already achieves state-of-the-art performance on the NAVSIM benchmark. Incorporating CoT-based guidance provides no gain, mainly due to the limited diversity of NAVSIM scenarios. Moreover, Bench2Drive already supplies explicit navigation instructions. For efficiency, we therefore keep VLM-based textual guidance optional.

\smallskip

\bigskip
\noindent\textbf{Q3.} \textit{Why was the model trained on datasets such as CODA-LM and OmniDrive, yet not evaluated on them?}

Our model is trained on a mixture of 12 open-source driving datasets, which enables handling multiple scenes and tasks. We observed that most existing benchmarks, including CODA-LM and OmniDrive, employ traditional metrics such as BLEU and ROUGE. These metrics evaluate performance by calculating the textual similarity between a model's response and the ground-truth answer. Such metrics primarily assess dataset-specific fitting rather than general reasoning ability. Therefore, to better evaluate the model's true Visual Question Answering (VQA) capabilities, we utilize the GPT-Score from the DriveLM and DriveBench benchmarks, as presented in Tab.~\ref{tab:drive_results}.

\smallskip

\bigskip
\noindent\textbf{Q4.} \textit{Does DriveVQA accurately reflect the capabilities of a VLM for autonomous driving?}

DriveBench has shown that many VLMs' decision accuracy does not degrade with visual quality, indicating a reliance on priors rather than genuine visual understanding. Therefore, we primarily use the Planning task to evaluate a model's practical capabilities in dynamic driving scenarios. However, for completeness and to facilitate comparison with existing work, we also provide the VQA results on the DriveLM and DriveBench benchmarks.
\smallskip

\bigskip
\noindent\textbf{Q5.} \textit{Does ReCogDrive rely on high-quality QA annotations to generalize to other domains?}

ReCogDrive achieves state-of-the-art results by leveraging high-quality driving QA datasets. However, a key contribution of our work is the scalable hierarchical data generation pipeline. Built on advanced VLMs deployed with Sglang, this pipeline can automatically generate large-scale QA annotations, followed by normalization, scoring, and filtering to ensure quality. This design ensures efficiency and adaptability, enabling transfer to new domains without costly manual annotation.

\smallskip

\bigskip
\noindent\textbf{Q6.} \textit{Why does applying the formula in Eq.~\ref{eq:pdms} to the averaged sub-metrics in Tab.~\ref{tab:comparison_modified} not yield the reported PDM Score?}

As documented in a NAVSIM issue~\footnote{\url{https://github.com/autonomousvision/navsim/issues/116}}, the PDMS score is calculated per scene using the formula and is then averaged across all scenes. Applying the formula to sub-metrics that have already been averaged is a different computation and will thus lead to a different final value.
\smallskip

\bigskip
\noindent\textbf{Q7.} \textit{Why does the paper adopt such a low discount factor (0.6) for a long-horizon driving task?}

In Eq.~\ref{eq:grpo}, the discount factor is applied across denoising steps rather than over the temporal horizon of the driving task. In diffusion-based planning, early denoising steps amplify high-variance noise and often destabilize learning. A smaller discount factor places greater weight on later, more reliable denoising steps, which directly improves stability and feasibility of generated trajectories. This design choice is thus motivated by the dynamics of the denoising process rather than by a short-horizon assumption of the driving task itself.

\smallskip

\bigskip
\noindent\textbf{Q8.} \textit{Why does the paper employ a BC loss instead of the standard KL-divergence loss?}

We employ a BC-style objective as a stable surrogate for KL regularization. Direct KL constraints often lead to unstable optimization in diffusion-based RL, since variance amplification across denoising steps can cause noisy and unreliable gradients. In contrast, the BC formulation provides smoother updates and better training stability, which proved crucial for policy learning effectiveness.

\smallskip

\bigskip
\noindent\textbf{Q9.} \textit{Does the diffusion planner only provide performance gains?}

As shown in \cref{tab:vlm_trajectory}, beyond the $+2.4$ PDMS improvement over plain-text VLM outputs, the diffusion planner also achieves $7.8\times$ faster inference by requiring only a forward pass rather than auto-regressive generation. More importantly, it eliminates the $0.01\%$ template mismatch errors observed in plain-text decoding, an unacceptable rate in safety-critical driving, making it both more accurate and more reliable for practical autonomous driving.

\section{More Results}
\label{sup:results}

\begin{table}[ht]
  \centering
  \caption{\textbf{Performance comparison on Navtest Benchmark with extended metrics.}}
  \label{tab:exp_results}
  \small
  \scalebox{0.85}{
  \begin{tabular}{l|ccccccccc|c}
    \toprule
    Method              & NC$\uparrow$ & DAC$\uparrow$ & EP$\uparrow$ & TTC$\uparrow$ & C$\uparrow$ & TL$\uparrow$ & DDC$\uparrow$ & LK$\uparrow$ & EC$\uparrow$ & EPDMS$\uparrow$ \\
    \midrule
    Transfuser~\citep{chitta2022transfuser}    & 97.7 & 92.8 & 79.2 & 92.8 & \textbf{100}   & 99.9 & 98.3 & 67.6 & 95.3 & 77.8 \\
    VADv2~\citep{chen2024vadv2}         & 97.3 & 91.7 & 77.6 & 92.7 & \textbf{100}   & 99.9 & 98.2 & 66.0 & 97.4 & 76.6 \\
    Hydra-MDP~\citep{li2024hydra}     & 97.5 & 96.3 & 80.1 & 93.0 & \textbf{100}   & 99.9 & 98.3 & 65.5 & 97.4 & 79.8 \\
    Hydra-MDP++~\citep{li2024hydra}   & 97.9 & \textbf{96.5} & 79.2 & 93.4 & \textbf{100}   & \textbf{100.0} & 98.9 & 67.2 & \textbf{97.7} & 80.6 \\
    ARTEMIS~\citep{feng2025artemis}           & 98.3 & 95.1 & 81.5 & 97.4 & \textbf{100} & 99.8 & 98.6 & 96.5 & 98.3 & 83.1 \\
    \midrule
    \textbf{ReCogDrive}  & \textbf{98.3} & 95.2 & \textbf{87.1} & \textbf{97.5} & 98.3 & 99.8 & \textbf{99.5} & \textbf{96.6} & 86.5 & \textbf{83.6} \\
    \bottomrule
  \end{tabular}}
\end{table}

\noindent\textbf{Experiments on NAVSIM with extended metrics.} Hydra MDP++~\citep{li2024hydra} introduces additional evaluation metrics: Traffic Light Compliance (TL), Lane Keeping Ability (LK), Driving Direction Compliance (DDC) and Extended Comfort (EC) to more comprehensively assess driving performance. We evaluate \name{} on NAVSIM using these extended metrics as well. Tab.~\ref{tab:exp_results} compares our approach against existing methods under this metrics. \name{} achieves state of the art scores in Driving Direction Compliance (DDC), Lane Keeping Ability (LK), Ego Progress (EP) and Time to Collision (TTC), and delivers a 0.5 improvement in EPDMS over ARTEMIS~\citep{feng2025artemis}, demonstrating the effectiveness of our method.

\noindent\textbf{Evaluations of VLMs across driving tasks.}
Tab.~\ref{tab:benchmark} reports results on \textit{DriveBench} across perception, prediction, planning, and behavior. ReCogDrive-VLM achieves the best overall performance and shows strong robustness under both corrupted and text-only settings, outperforming general-purpose VLMs (e.g., Qwen2-VL, InternVL2) and specialist models (e.g., DriveLM, Dolphins). These results demonstrate the strong driving scene understanding capability of our ReCogDrive-VLM.

\begin{table*}[t]
    \centering
    \caption{\textbf{Evaluation of ReCogDrive-VLM on DriveBench.} 
    We report ReCogDrive-VLM’s performance across four key driving dimensions: perception, prediction, planning, and behavior. Results are shown under three input settings: \textcolor{robo_green}{Clean} (unaltered images), \textcolor{robo_red}{Corr.} (images with averaged corruptions), and \textcolor{robo_blue}{T.O.} (text-only inputs). }
    \vspace{-0.2cm}
    \label{tab:benchmark}
    \resizebox{\linewidth}{!}{
    \begin{tabular}{r|r|c|ccc|ccc|ccc|ccc}
    \toprule
    \multirow{2}{*}{\textbf{Method}} & \multirow{2}{*}{\textbf{Size}} & \multirow{2}{*}{\textbf{Type}} & \multicolumn{3}{c|}{\includegraphics[width=0.024\linewidth]{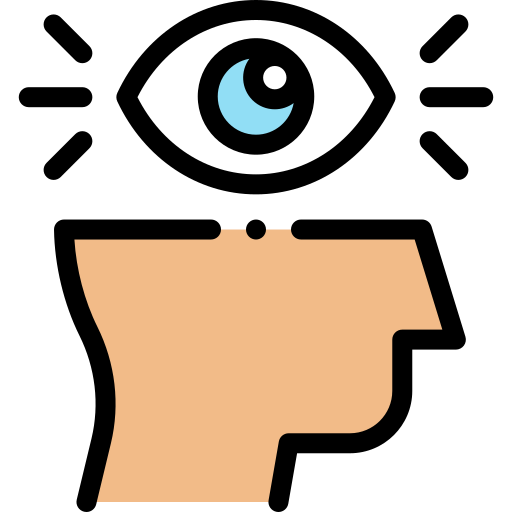}~\textbf{Perception}} & \multicolumn{3}{c|}{\includegraphics[width=0.024\linewidth]{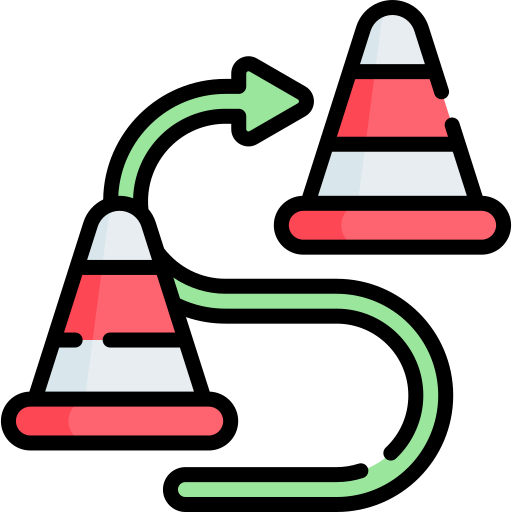}~\textbf{Prediction}} & \multicolumn{3}{c|}{\includegraphics[width=0.024\linewidth]{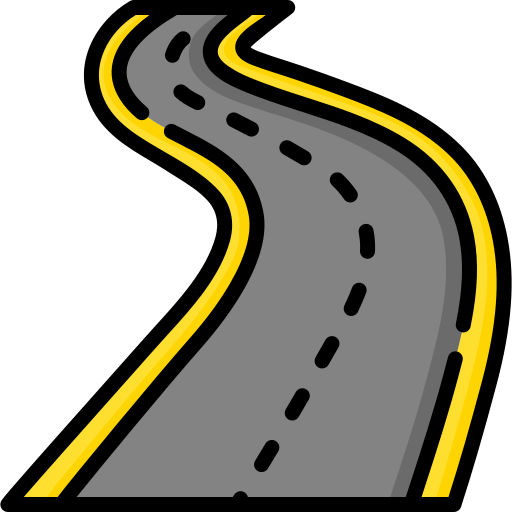}~\textbf{Planning}} & \multicolumn{3}{c}{\includegraphics[width=0.024\linewidth]{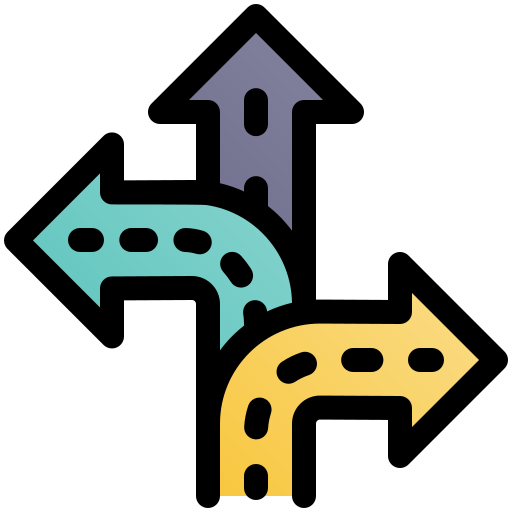}~\textbf{Behavior}}
    \\
    & & & \textcolor{robo_green}{{Clean}} & \textcolor{robo_red}{{Corr.}} & \textcolor{robo_blue}{{T.O.}} & \textcolor{robo_green}{{Clean}} & \textcolor{robo_red}{{Corr.}} & \textcolor{robo_blue}{{T.O.}} & \textcolor{robo_green}{{Clean}} & \textcolor{robo_red}{{Corr.}} & \textcolor{robo_blue}{{T.O.}} & \textcolor{robo_green}{{Clean}} & \textcolor{robo_red}{{Corr.}} & \textcolor{robo_blue}{{T.O.}}
    \\\midrule\midrule
    \cellcolor{robo_green!10}\includegraphics[width=0.024\linewidth]{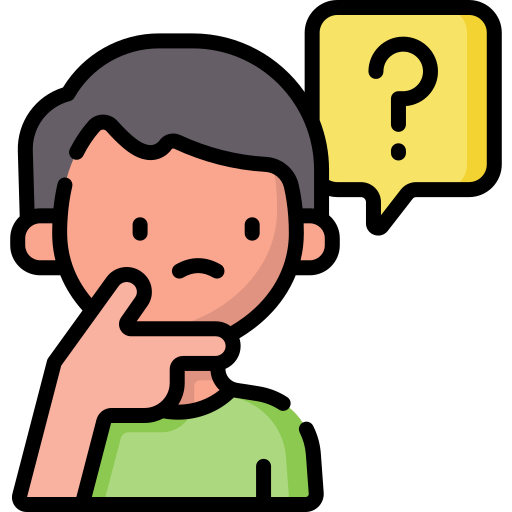}~\textcolor{robo_green}{Human} & \cellcolor{robo_green!10}\textcolor{robo_green}{-} & \cellcolor{robo_green!10}\textcolor{robo_green}{-} & \cellcolor{robo_green!10}\textcolor{robo_green}{$47.67$} & \textcolor{robo_red}{$38.32$} \cellcolor{robo_green!10} & \cellcolor{robo_green!10}\textcolor{robo_green}{-} & \cellcolor{robo_green!10}\textcolor{robo_green}{-} & \cellcolor{robo_green!10}\textcolor{robo_green}{-} & \cellcolor{robo_green!10}\textcolor{robo_green}{-} & \cellcolor{robo_green!10}\textcolor{robo_green}{-} & \cellcolor{robo_green!10}\textcolor{robo_green}{-} & \cellcolor{robo_green!10}\textcolor{robo_green}{-} & \cellcolor{robo_green!10} \textcolor{robo_green}{$69.51$} & \cellcolor{robo_green!10} \textcolor{robo_red}{$54.09$} & \cellcolor{robo_green!10}\textcolor{robo_green}{-}
    \\\midrule
    \textcolor{gray}{GPT-4o~\citep{hurst2024gpt}} & - & \textcolor{gray}{Commercial} & \textcolor{gray}{$35.37$} & \textcolor{gray}{$35.25$} &  \textcolor{gray}{$36.48 $} & \textcolor{gray}{$51.30$} & \textcolor{gray}{$49.94$} & \textcolor{gray}{$49.05$} & \textcolor{gray}{$75.75$} & \textcolor{gray}{$75.36$} & \textcolor{gray}{$73.21$} & \textcolor{gray}{$45.40$} & \textcolor{gray}{$44.33$} &  \textcolor{gray}{$50.03$} 
    \\\midrule
    LLaVA-1.5~\citep{liu2023visual} & $7$ B & Open & $23.22$ & $22.95$ & $22.31$ & $22.02$ & $17.54$ & $14.64$ & $29.15$ &  $31.51$ &  $32.45$ & $13.60$ &  $13.62$  & $14.91$
    \\
    LLaVA-1.5~\citep{liu2023visual} & $13$ B & Open & $23.35$ &  $23.37$ & $22.37$ & $36.98$ &  $37.78$ & $23.98$ & $34.26$ &  $34.99$ &  $38.85$ & $32.99$ & $32.43$ & $32.79$
    \\
    LLaVA-NeXT~\citep{liu2024llavanext} & $7$ B & Open & $24.15$ & $19.62$ & $13.86$ & $35.07$ &  $35.89$ & $28.36$ & $45.27$ & $44.36$ & $27.58$ & $48.16$ & $39.44$ & $11.92$
    \\
    InternVL2~\citep{chen2024far} & $8$ B & Open & $\underline{32.36}$ &  $\underline{32.68}$ &  $33.60$ & $45.52$ & $37.93$ &  $\underline{48.89}$ & $53.27$ &  $55.25$ & $34.56$ &  $\mathbf{54.58}$ & $40.78$ & $20.14$ 
    \\
    Phi-3~\citep{abdin2024phi} & $4.2$ B & Open & $22.88$ &  $23.93$ &  $28.26$ & $40.11$ & $37.27$ & $22.61$ & $60.03$ &  $61.31$ & $46.88$ & $45.20$ & $44.57$ & $28.22$
    \\
    Phi-3.5~\citep{abdin2024phi} & $4.2$ B & Open & $27.52$ & $27.51$ &  $28.26$ & $45.13$ & $38.21$ & $4.92$ & $31.91$ & $28.36$ &  $46.30$ & $37.89$ &  $\underline{49.13}$  & $39.16$
    \\
    Oryx~\citep{liu2024oryx} & $7$ B & Open & $17.02$ &  $15.97$ &  $18.47$ & $48.13$ & $\underline{46.63}$ & $12.77$ & $53.57$ &  $55.76$ & $48.26$ & $33.92$ & $33.81$ & $23.94$
    \\
    Qwen2-VL~\citep{Qwen2-VL} & $7$ B & Open & $28.99$ & $27.85$ &  $\underline{35.16}$ & $37.89$ &  $39.55$ & $37.77$ & $57.04$ & $54.78$ & $41.66$ & $49.07$ &  $47.68$ &  $\mathbf{54.48}$
    \\
    Qwen2-VL~\citep{Qwen2-VL} & $72$ B & Open & $30.13$ & $26.92$ & $17.70$ & $\mathbf{49.35} $ & $43.49$ & $5.57$ & $61.30$ &  $63.07$ & $53.35$ & $\underline{51.26}$ & $\mathbf{49.78}$ & $39.46$
    \\
    \midrule
    DriveLM~\citep{sima2024drivelm} & $7$ B & Specialist & $16.85$ & $16.00$ & $8.75$ & $44.33$ & $39.71$ & $4.70$ & $\underline{68.71}$ & $\underline{67.60}$ & $\underline{65.24}$ & $42.78$ & $40.37$ & $27.83$ 
    \\
    Dolphins~\citep{ma2024dolphins} & $7$ B & Specialist & $9.59$ &  $10.84$ &  $11.01$ & $32.66$ & $29.88$ &  $39.98$ & $52.91$ &  $53.77$ &  $60.98$ & $8.81$ & $8.25$ & $11.92$
    \\
    ReCogDrive-VLM & $8$ B & Specialist & $\mathbf{64.95}$ &  $\mathbf{63.35}$ &  $\mathbf{71.44}$ & $\underline{49.34}$ & $\mathbf{48.92}$ &  $\mathbf{53.49}$ & $\mathbf{70.20}$ &  $\mathbf{72..5}$ &  $\mathbf{71.76}$ & $42.36$ & $42.76$ & $\underline{44.00}$
    \\
    \bottomrule
\end{tabular}}
\end{table*}

\begin{table*}[t!]
    \centering
    \hfill
    \begin{minipage}[t]{0.5\linewidth}
        \centering
        \caption{Impact of Different BC Loss Weights $\lambda$.}
        \renewcommand\arraystretch{1.1}
        \renewcommand\tabcolsep{3.5pt}
        \small
        \scalebox{0.96}{
        \begin{tabular}{c|ccccc|c}
            \toprule
            BC Wt. & NC & DAC & TTC & Conf. & EP & \cellcolor{gray!30}PDMS \\
            \midrule
            0.001 & 97.9 & 97.3 & 94.9 & 100.0 & 87.3 & \cellcolor{gray!30}90.8 \\
            0.01  & 97.9 & 97.5 & 94.9 & 100.0 & 86.0 & \cellcolor{gray!30}90.3 \\
            0.1  & 97.6 & 97.2 & 93.9 & 100.0 & 85.0 & \cellcolor{gray!30}89.3 \\
            \bottomrule
        \end{tabular}}
        % \caption{Impact of Different BC Loss Weights $\lambda$.}
        \label{tab:bcloss}
    \end{minipage}%
    \hfill
    \begin{minipage}[t]{0.5\linewidth}
        \centering
        \caption{Impact of Different Min Samplings $\sigma_{\min}^{\exp}$.}
        \renewcommand\arraystretch{1.1}
        \renewcommand\tabcolsep{3.5pt}
        \small
        \scalebox{0.96}{
        \begin{tabular}{c|ccccc|c}
            \toprule
            Min Samp.  & NC & DAC & TTC & Conf. & EP & \cellcolor{gray!30}PDMS \\
            \midrule
            0.01 & 98.2 & 97.8 & 95.6 & 100.0 & 82.8 & \cellcolor{gray!30}89.5 \\
            0.02 & 97.9 & 97.3 & 94.9 & 100.0 & 87.3 & \cellcolor{gray!30}90.8 \\
            0.04 & 97.9 & 97.7 & 95.3 & 100.0 & 85.7 & \cellcolor{gray!30}90.5 \\
            \bottomrule
        \end{tabular}}
        % \caption{Impact of Different Min Samplings $\sigma_{\min}^{\exp}$.}
        \label{tab:mixrate}
    \end{minipage}
\end{table*}

\begin{table*}[t!]
  \centering
  \begin{minipage}[t]{0.48\linewidth}
    \centering
    \caption{Comparison between trajectory-only (Only Traj), adding high-level command (w/ High Com.), and chain-of-thought (w/ CoT).}
    \footnotesize
    \setlength{\tabcolsep}{6pt}
    \renewcommand\arraystretch{1.1}
    \renewcommand\tabcolsep{3.5pt}
    \scalebox{0.92}[0.96]{
    \begin{tabular}{c| c c c c c | c}
      \toprule
      \makecell{Mode} & NC & DAC & TTC & Conf. & EP & \cellcolor{gray!30}PDMS \\
      \midrule
      Only Traj & 97.9 & 97.3 & 94.9 & 100.0 & 87.3 & \cellcolor{gray!30}90.8 \\
      w/ High Com. & 97.9 & 97.3 & 94.9 & 100.0 & 87.2 & \cellcolor{gray!30}90.8 \\
      w/ CoT & 97.8 & 97.2 & 94.8 & 100.0 & 87.1 & \cellcolor{gray!30}90.7 \\
      \bottomrule
    \end{tabular}}
    \label{tab:traj_com}
\end{minipage}
  \hfill
  \begin{minipage}[t]{0.48\linewidth}
    \centering
    \caption{Impact of different discount factors $\gamma$. We use discounting to reduce the influence of early-step noise during the diffusion process.}
    \renewcommand\arraystretch{1.1}
    \renewcommand\tabcolsep{3.5pt}
    \footnotesize
    \scalebox{1.0}[0.96]{
    \begin{tabular}{c|ccccc|c}
        \toprule
        Discount  & NC & DAC & TTC & Conf. & EP & \cellcolor{gray!30}PDMS \\
        \midrule
        0.6 & 97.9 & 97.3 & 94.9 & 100.0 & 87.3 & \cellcolor{gray!30}90.8 \\
        0.8 & 98.0 & 97.6 & 95.3 & 100.0 & 85.9 & \cellcolor{gray!30}90.5 \\
        1.0 & 97.9 & 97.5 & 94.9 & 100.0 & 86.0 & \cellcolor{gray!30}90.3 \\
        \bottomrule
    \end{tabular}}
    % \caption{Impact of different discount factors $\gamma$. We use discounting to reduce the influence of early-step noise during the diffusion process.}
    \label{tab:discount}
  \end{minipage}
  \vspace{-0.6cm}
\end{table*}

% \paragraph{Scaling Laws of QA Data for Planning Performance} 

% We collected a large-scale, high-quality driving QA dataset to adapt VLMs to real-world driving scenarios. As shown in Tab.~\ref{tab:training-qa-growth}, increasing the number of QA pairs from 85k to 3.2M increases the PDMS from 83.3 to 85.3, demonstrating the necessity of driving-specific pre-training and confirming that scaling laws hold under our conditions. Furthermore, filtering and rewriting the entire 3.2M set yields an additional 0.9 PDMS improvement to 86.2, highlighting the critical importance of data quality.

\paragraph{Impact of Different Behavior Clone Loss Weights.} 
Tab.~\ref{tab:bcloss} examines the impact of the BC loss weight~$\lambda$. As $\lambda$ decreases, the policy learns more aggressively, and at $\lambda=0.001$ ego progress increases to $87.3$ with slight declines in NC, DAC, and TTC, yielding the highest PDMS.

\paragraph{Impact of Different Min Sampling Values.} 
Following~\citep{ren2024diffusion,nichol2021improved}, we apply a cosine schedule to the diffusion noise variances $\sigma_k$ and clip them to a minimum value $\sigma_{\min}^{\exp}$. Clipping $\sigma_k$ to a nonzero minimum encourages the diffusion policy to sample more diverse trajectories, yet overly large $\sigma_{\min}^{\exp}$ can destabilize training. Setting $\sigma_{\min}^{\exp}=0.02$ achieves a PDMS of 90.8, as shown in Tab.~\ref{tab:mixrate}.

\paragraph{Effect of VLM guidance modes.} 
As shown in Tab.~\ref{tab:traj_com}, outputting only trajectory achieves performance comparable to adding high-level command guidance, with almost identical PDMS scores. Interestingly, incorporating chain-of-thought reasoning does not further improve the results; instead, it slightly decreases the PDMS score by $0.1$. This suggests that the current NAVSIM benchmark lacks sufficient diversity to demonstrate the benefits of high-level command and CoT reasoning, so we keep these guidance optional.

\paragraph{Impact of Different Discount Factors.}
Tab.~\ref{tab:discount} shows the impact of the discount factor $\gamma$ on RL fine-tuning. When $\gamma=1.0$, all timesteps, including the very noisy early steps, contribute equally to the policy gradient, which amplifies high variability noise and destabilizes learning. In contrast, setting $\gamma=0.6$ focuses the update on later, more reliable steps and achieves the best PDMS of 90.8.

\section{Training Datasets Construction}
\label{sup:dataset_construction}

\subsection{Data Collection}
We collect 12 open-source driving QA datasets, including Talk2Car~\citep{deruyttere2019talk2car}, SUTD~\citep{xu2021sutd}, NuScenes-QA~\citep{qian2024nuscenes}, OmniDrive~\citep{wang2024omnidrive}, and others, yielding over 3.1 million question-answer pairs that cover perception, prediction, and planning tasks across diverse real-world scenarios.

\noindent\textbf{Talk2car}~\citep{deruyttere2019talk2car}  is built on the nuScenes~\citep{caesar2020nuscenes} dataset and contains 850 videos from the nuScenes~\citep{caesar2020nuscenes} training set. This dataset has 11,959 natural language commands.

\noindent\textbf{SUTD}~\citep{xu2021sutd} contains 10,080 in-the-wild videos and annotated 62,535 QA pairs. These videos are obtained through a combination of online collection and offline shooting, covering various weather conditions, times, and road conditions.

\noindent\textbf{DRAMA}~\citep{malla2023drama} is a dataset collected to investigate risk location and natural language description in driving scenarios. It contains 17,785 interactive driving scenarios.

\noindent\textbf{NuScenes-QA}~\citep{qian2024nuscenes} encompasses 34,149 complex autonomous driving scenes and 459,941 question-answer pairs, with various types of questions. It aims to evaluate a model's ability to understand and reason about complex visual data in multi-modal, multi-frame, and outdoor scenarios.

\noindent\textbf{DriveGPT4}~\citep{xu2024drivegpt4} is built based on the BDD-X~\citep{kim2018textual} dataset, which contains about 20,000 samples. By dividing them into 16,803 training segments and 2,123 testing segments, question-answer pairs are generated using the control signal data and text annotations. 

\noindent\textbf{LingoQA}~\citep{marcu2024lingoqa} contains 28K unique short video scenarios and 419K annotations. The dataset covers various questions related to driving scenarios, including aspects such as behaviors, scenery, and perception.

\noindent\textbf{DriveLM}~\citep{sima2024drivelm} consists of a training set of 4,072 frames and a validation set of 799 frames, with an average of 91.4 QA pairs per frame. 

\noindent\textbf{MAPLM}~\citep{cao2024maplm} contains point-cloud BEV projections and surround view images of various traffic scenes, such as highways and urban roads, and is equipped with element-level, lane-level, and road-level scene descriptions.

\noindent\textbf{NuInstruct}~\citep{ding2024holistic} is a dataset constructed based on Nuscenes~\citep{caesar2020nuscenes}, containing 91K multi-view video instruction-response pairs in 17 subtasks.

\noindent\textbf{CODA-LM}~\citep{chen2024automated} comprises 9,768 real-world driving scenarios with 41,722 textual annotations for critical road entities and 21,537 annotations for road corner cases.

\noindent\textbf{OminiDrive}~\citep{wang2024omnidrive} covers 3D perception, reasoning, and planning tasks, including offline and online question-response tasks.

\noindent\textbf{Senna}~\citep{jiang2024senna} design a series of planning-oriented QAs including scene description, traffic signal detection, vulnerable road user identification, motion intention prediction, meta-action planning and planning explanation. Since the senna dataset is not publicly available, we used Qwen2.5-VL-72B~\citep{Qwen2.5-VL} for question-answer data annotation.

% \begin{figure}[!t]
%     \centering
%     \includegraphics[width=1.0\linewidth]{sec/figures/ICLR-supp1.pdf}
%     \vspace{-10pt}
%     \caption{\textbf{Dataset Construction Pipeline.} Step 1: collect open-source driving QA datasets and NAVSIM samples. Step 2: process data via normalization, augmentation, and filtering. Step 3: automatic annotation pipeline generates QA pairs for perception, prediction, and planning tasks. The pie chart summarizes category distribution.}
%     \label{fig:intro}
%     \vspace{-10pt}
% \end{figure}

\subsection{Scalable Hierarchical Data Pipeline}
To systematically enhance VLM performance for autonomous driving, we design a scalable hierarchical data pipeline consisting of three stages: \textit{Generation}, \textit{Refinement}, and \textit{Quality Control}, as shown in Fig.~\ref{fig:representation_gen}. This design allows us to construct diverse, semantically consistent, and high-quality annotations at scale.  

\noindent\textbf{Generation.}  
We first leverage Qwen2.5-VL~\citep{Qwen2.5-VL} with crafted prompts to generate annotations across the full spectrum of autonomous driving tasks on NAVSIM~\citep{dauner2025navsim}. These tasks span perception (e.g., scene description, key object identification, road marking recognition, traffic light classification, vulnerable road user detection), prediction (e.g., motion prediction), planning (e.g., high-level command prediction and decision explanation), and advanced reasoning (e.g., counterfactual scenarios inspired by OmniDrive~\citep{wang2024omnidrive}). To strengthen supervision, we further incorporate NAVSIM’s existing sensor and trajectory labels into the generation process. Applying this stage yields a large pool of candidate QA pairs.  

\noindent\textbf{Refinement.} Since open-source datasets employ heterogeneous formats, we unify them through normalization and linguistic enrichment.  
(i) \textit{Data Normalization:} Different datasets adopt different bounding-box styles. For example, DriveLM~\citep{sima2024drivelm} represents an object as ``\textless c2,CAM\_BACK,864.2,468.3\textgreater'', while NuInstruct~\citep{ding2024holistic} uses ``\textless car\textgreater [c2,584,478,603,516]''. We convert all such variants into a standardized tag format:  
``\textless car\textgreater \textless FRONT\_VIEW\textgreater \textless box\textgreater [$x_1, y_1, x_2, y_2$]\textless /box\textgreater''  
following the InternVL3~\citep{zhu2025internvl3} pre-training convention. Coordinates \(x_1, y_1, x_2, y_2\) are rescaled to the integer range [0,1000] relative to image resolution.  
(ii) \textit{Data Augmentation:} To enrich linguistic diversity, especially for datasets with limited QA templates (e.g., CODA-LM with only three formats, DRAMA with terse answers), we employ a two-step process. First, an LLM paraphrases question templates to expand structural variation. Second, Qwen2.5-VL is prompted with the original image, question, and answer to generate more detailed, natural responses. This refinement ensures consistency in structure and greater variety in language.  

\noindent\textbf{Quality Control.}  
Finally, we implement a rigorous filtering step. Qwen2.5-VL assigns each QA pair a quality score based on predefined linguistic and semantic criteria. Any pair scoring below 60 is discarded. After this filtering process, we retain 2.3M high-quality QA pairs. Notably, applying the pipeline to NAVSIM produces 775K well-structured pairs, which we use to fine-tune the VLM, substantially enhancing its planning and reasoning capabilities.

\subsection{Dataset Statistics} 
The dataset consists of 41\% perception samples, 11\% prediction samples, and 24\% planning samples, aimed at improving VLMs’ understanding of driving scenarios. We also include 24\% visual instruction samples to maintain the model’s instruction‐following ability.

\section{Experiment Details}
\label{sup:implementation_details}
In this section, we first introduce the evaluation metrics of NAVSIM and Bench2Drive, and then detail our model configuration, the hyperparameters used in the three-stage training and evaluation, and our hardware setup.

\noindent\textbf{Evaluation Metrics.}
We evaluate our method on two primary benchmarks. For the planning-oriented NAVSIM benchmark, we use the official Predictive Driver Model Score (PDMS), a closed-loop metric that holistically assesses safety (NC, DAC, TTC), comfort (C), and progress (EP):
\begin{equation}
    \label{eq:pdms}
    \mathrm{PDMS} = \mathrm{NC} \times \mathrm{DAC} \times \left( \frac{5 \cdot \mathrm{EP} + 5 \cdot \mathrm{TTC} + 2 \cdot \mathrm{C}}{12} \right).
\end{equation}

For the CARLA-based Bench2Drive benchmark, which focuses on safety-critical scenarios, we report the Success Rate (SR) and the average Driving Score (DS). The DS combines Route Completion (RC) with multiplicative penalties for infractions (IS):
\begin{equation}
    \label{eq:b2d_metrics}
    \mathrm{SR} = \frac{N_{\text{success}}}{N_{\text{total}}}; \quad \mathrm{DS} = \frac{1}{N_{\text{total}}} \sum_{i=1}^{N_{\text{total}}} \left( \mathrm{RC}_i \times \prod_{j} \mathrm{IS}_{i,j} \right).
\end{equation}

\noindent\textbf{Model Architecture and Hyperparameter Details.}
Our model architecture consists of two main components: Vision-Language Models (VLMs) and a diffusion-based trajectory planner.
For the Vision-Language Models, we utilize the pre-trained InternVL3-8B and InternVL3-2B~\citep{zhu2025internvl3} model. The model processes visual inputs by splitting the image into patches, with a maximum of 12 dynamic patches allowed. The input images are cropped into 448x448 pixel patches.
For the diffusion planner, we use the DiT~\citep{peebles2023scalable} architecture, employing DDPM/DDIM~\citep{ho2020denoising} denoising methods. During training, the denoising process involves 100 steps, while for inference, the diffusion process is reduced to 5 steps. The hidden layer size of the DiT architecture is set to 384, with a head size of 8 and 16 layers.

\noindent\textbf{Training Configuration.} In the first stage, we fine-tune the VLM on the combined driving QA dataset for three epochs, with all parameters unfrozen and a batch size of 1,024. We use AdamW with a base learning rate of $4e^{-5}$, weight decay of 0.05, and a cosine learning rate schedule with a 10\% linear warmup. In the second stage, we train the diffusion-based planner with behavior cloning for 200 epochs and a batch size of 512. We use AdamW with a learning rate that warms up to $1e^{-4}$ over the first 1.5\% of steps, then decays cosinely to $1e^{-6}$, and a weight decay of $1e^{-4}$. In the third stage, we perform simulator-assisted reinforcement learning for 10 epochs with a batch size of 128, using AdamW and a cosine schedule that decays the learning rate from $1e^{-4}$ to 0.  Additionally, we stabilize diffusion training and inference by clipping the reconstructed clean estimate \(x_0\) to \(\pm 1.0\), clipping Gaussian noise samples to \(\pm 5.0\), enforcing a nonzero floor on the denoising standard deviation \(\sigma_t\) of 0.02 for training, clamping \(\sigma_t\) to at least 0.10 when computing \(\log p(x_{t-1}\!\mid\!x_t)\), and applying a discount factor \(\gamma=0.6\) during RL fine-tuning to downweight early, noisy timesteps.  
\begin{table}[ht]
  \centering
  \small
  \setlength{\tabcolsep}{6pt}
  \renewcommand{\arraystretch}{1.0}
  \caption{\textbf{Hyperparameters for \name{}.}}
  \begin{tabular}{c l l}
    \toprule
    \textbf{Stage} & \textbf{Parameter}         & \textbf{Value}                           \\
    \midrule
    \multirow{6}{*}{I}  
      & Number of epochs          & 3                                        \\
      & Batch size                & 1024                                  \\
      & Learning rate             & $4e^{-5}$                        \\
      & Weight decay              & 0.05                                     \\
      & Warmup ratio              & 0.10                                     \\
      & Learning rate schedule    & Cosine                                   \\
    \midrule
    \multirow{4}{*}{II} 
      & Number of epochs          & 200                                      \\
      & Batch size                & 512                                      \\
      & Learning rate             & $1e^{-4}$                         \\
      & Learning rate schedule    & Cosine                       \\
      & Weight decay              & $1\times10^{-4}$                         \\
    \midrule
    \multirow{11}{*}{III}
      & Number of epochs          & 10                                       \\
      & Batch size                & 128                                      \\
      & Learning rate             & $1e^{-4}$                          \\
      & Learning rate schedule    & Cosine                       \\
      & BC loss weight            & $1e^{-2}$              \\
      & Denoised clip threshold   & $\pm1.0$                                 \\
      & Noise clip threshold      & $\pm3.0$                              \\
      & Minimum denoising std     & 0.02                                \\
      & Minimum log‐variance std  & 0.10                                \\
      & Discount factor           & 0.6                                 \\
    \bottomrule
  \end{tabular}
  \label{tab:training-hyperparams}
\end{table}

\noindent\textbf{Hardware Configuration.} We implement \name{} on Debian with PyTorch, training across four nodes—each equipped with an Intel\textregistered{} Xeon\textregistered{} Platinum 8457C CPU and eight NVIDIA H20 GPUs (32 GPUs in total). Inference is performed on a single node with 8 GPUs.

% \begin{table}[h!]
%     \centering
%     \small
%     \begin{tabular}{c|ccc}
%     \hline
%         \multirow{2}{*}{Settings} & \multicolumn{3}{c}{\name{}} \\ 
%         & Stage 1 & Stage 2 & Stage 3 \\ 
%         \hline
%         Dataset             &  Percp.\&Pred.     & ALL     & Traj. \\ 
%         Trainable           & Full Model    & Full Model   & Full Model \\ 
%         Packed Batch Size   & 1024    & 1024      & 512 \\ 
%         Learning Rate       & 4e-5   & 1e-5      & 1e-6 \\ 
%         Context Length      & 4096  & 4096     & 4096 \\
%         LR Scheduler        & Constant   & Cosine      & Cosine \\ 
%         Training Epochs     & 3      & 2       & 1 \\ 
%         Training Tokens     & $\sim$31B & $\sim$146B & $\sim$44B \\ 
%     \end{tabular}
%     \caption{Training configurations for \name{}.}
% \end{table}

\section{Limitations}
\label{sup:limitations}
% Although \name{} achieves state of the art on NAVSIM, it directly employs a pre-trained Vision-Language Model without any architectural modifications. Designing a vision encoder capable of effectively processing three dimensional spatial imagery together with temporal sequences remains an open challenge. In addition, inference with large vision language models is computationally intensive, which limits deployment in real time systems. Finally, our current approach does not fuse lidar data and therefore cannot fully leverage precise geometric measurements. Addressing these issues will be important for future work.  
Although \name{} achieves state of the art performance on the NAVSIM benchmark, it still faces several limitations. These include the difficulty of processing multiple camera inputs, handling video frame sequences, and relatively high inference latency. Future work may address these issues by designing a 3D vision encoder that aligns with textual features and developing more efficient model architectures. We also plan to deploy and evaluate our model on real vehicles.  

\section{LLM Usage Statement}
\label{sup:llm_usage}
Large Language Models (LLMs) were employed solely as writing assistants to improve the clarity, grammar, and readability of the manuscript. They were not involved in research ideation, methodology design, data analysis, or result interpretation. 

\section{Broader Impacts}
\label{sup:broader_impacts}
Our research promotes the application of Vision-Language Models (VLMs) in the autonomous driving domain. Through simulator-assisted reinforcement learning, our model more effectively mitigates collision risk and generalizes to rare, challenging scenarios. This advancement could lead to safer, more reliable autonomous systems capable of handling real-world driving scenarios.

\section{Qualitative Results} 
\label{sup:qualitative} 
We first present extensive qualitative visualizations of \name{} on NAVSIM and Bench2Drive, highlighting its ability to produce safe and smooth trajectories. Next, we provide trajectory comparisons between ReCogDrive-IL and ReCogDrive-RL to demonstrate the effectiveness of our proposed DiffGRPO. We further showcase dialogue examples generated by ReCogDrive, illustrating its cognitive reasoning capability. Finally, we analyze representative failure cases on NAVSIM to provide insights into current limitations.  

\subsection{Qualitative Results on Navtest Splits}
\begin{figure*}[htbp]
\centering
\resizebox{0.97\linewidth}{!}{
\includegraphics[width=1.00\linewidth]{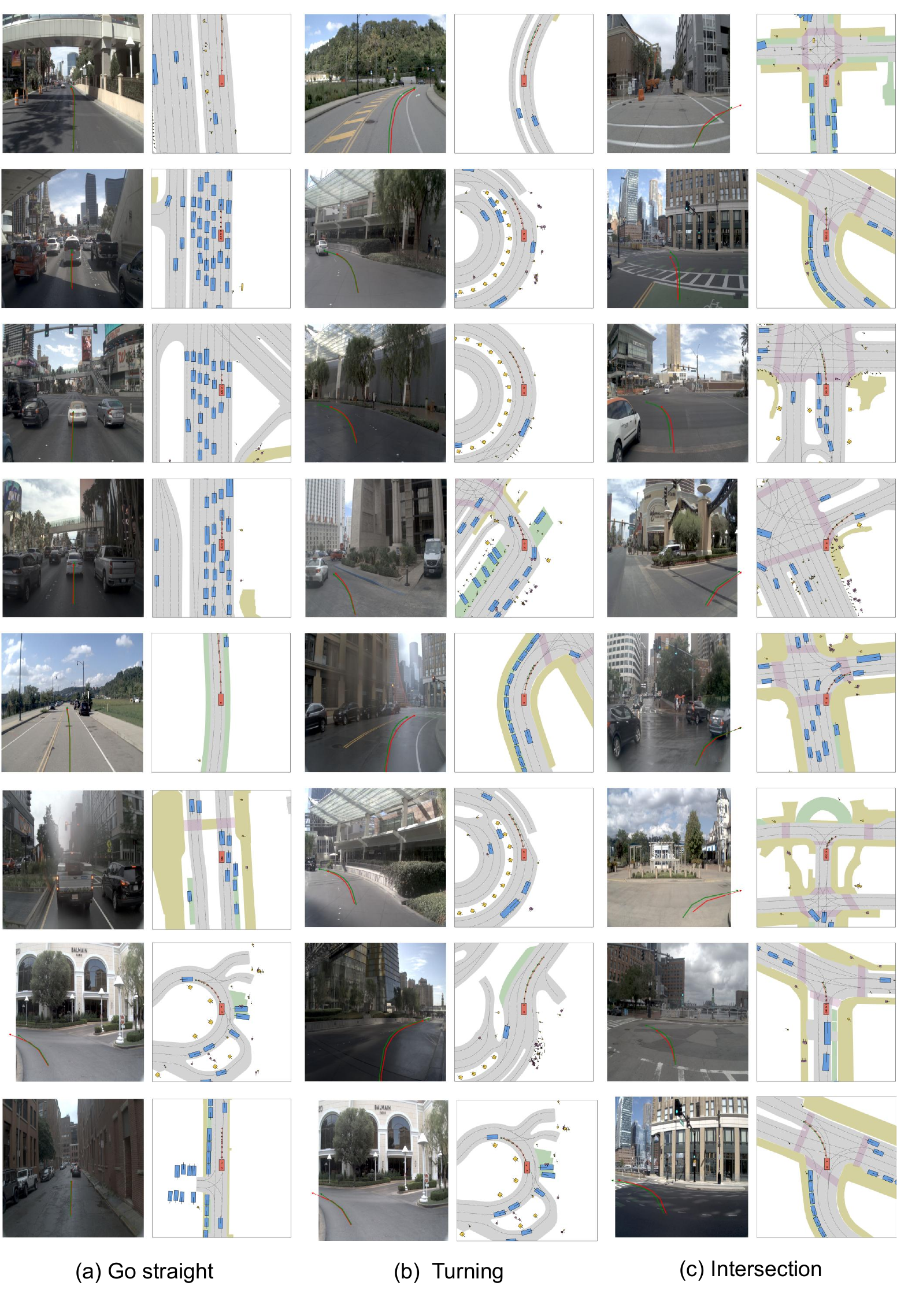}}
    \caption{Qualitative results on the Navtest benchmark.}
    \label{fig:nav0}
    \vspace{-5pt}
\end{figure*}

We present representative cases from the Navtest splits, highlighting \name{}'s ability to follow navigation instructions while ensuring safety and smoothness (see Fig.~\ref{fig:nav0} and Fig.~\ref{fig:nav1}).

\begin{figure*}[htbp]
\centering
\resizebox{0.97\linewidth}{!}{
\includegraphics[width=1.00\linewidth]{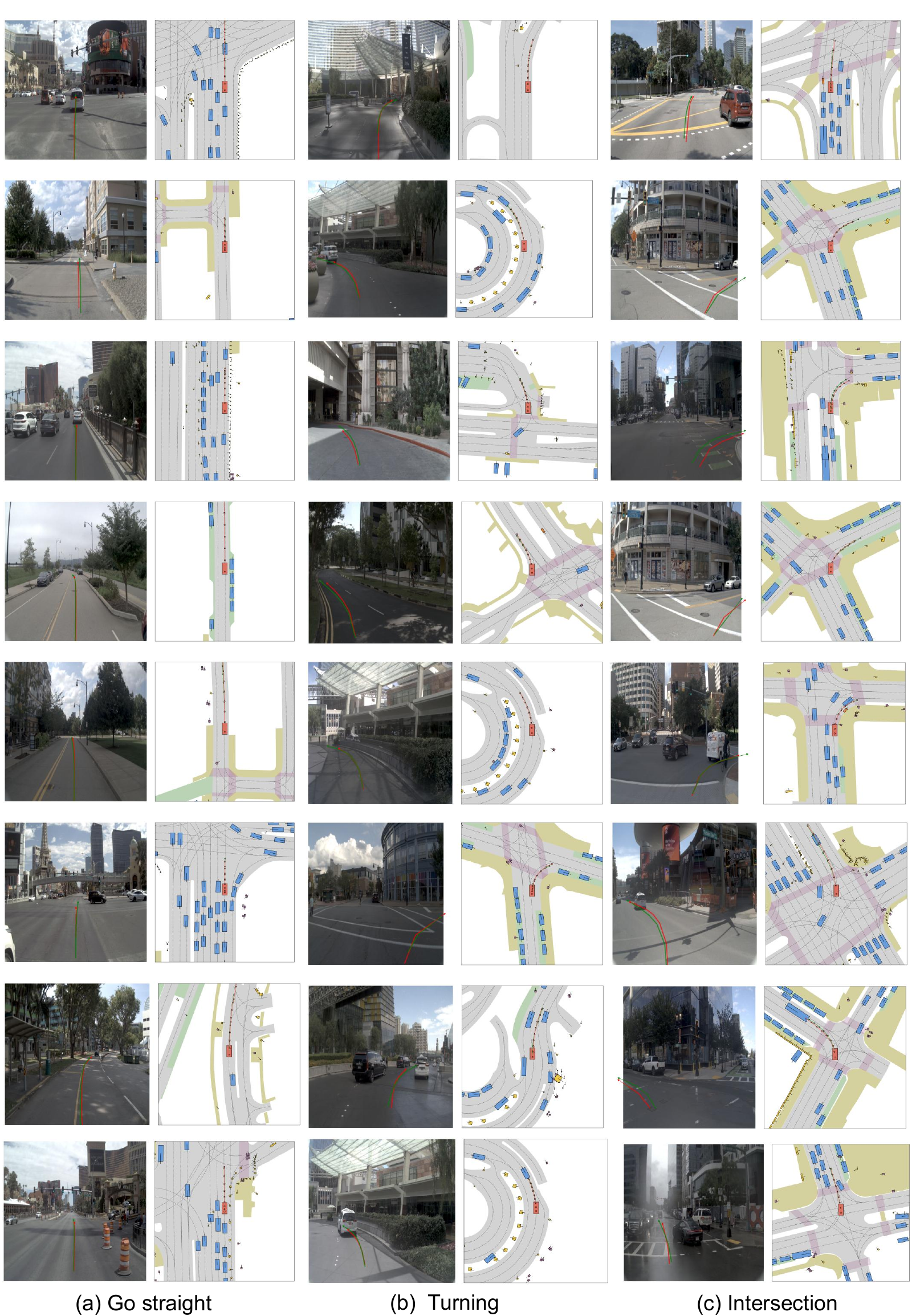}}
    \caption{Qualitative results on the Navtest benchmark.}
    \label{fig:nav1}
    \vspace{-5pt}
\end{figure*}

\subsection{Qualitative Results on Bench2Drive Test Scenarios}
We evaluate \name{} on challenging scenarios from Bench2Drive, demonstrating its robustness under diverse traffic conditions and rare corner cases, while consistently maintaining strong performance in closed-loop settings.

\begin{figure*}[htbp]
\centering
\resizebox{0.97\linewidth}{!}{
\includegraphics[width=1.00\linewidth]{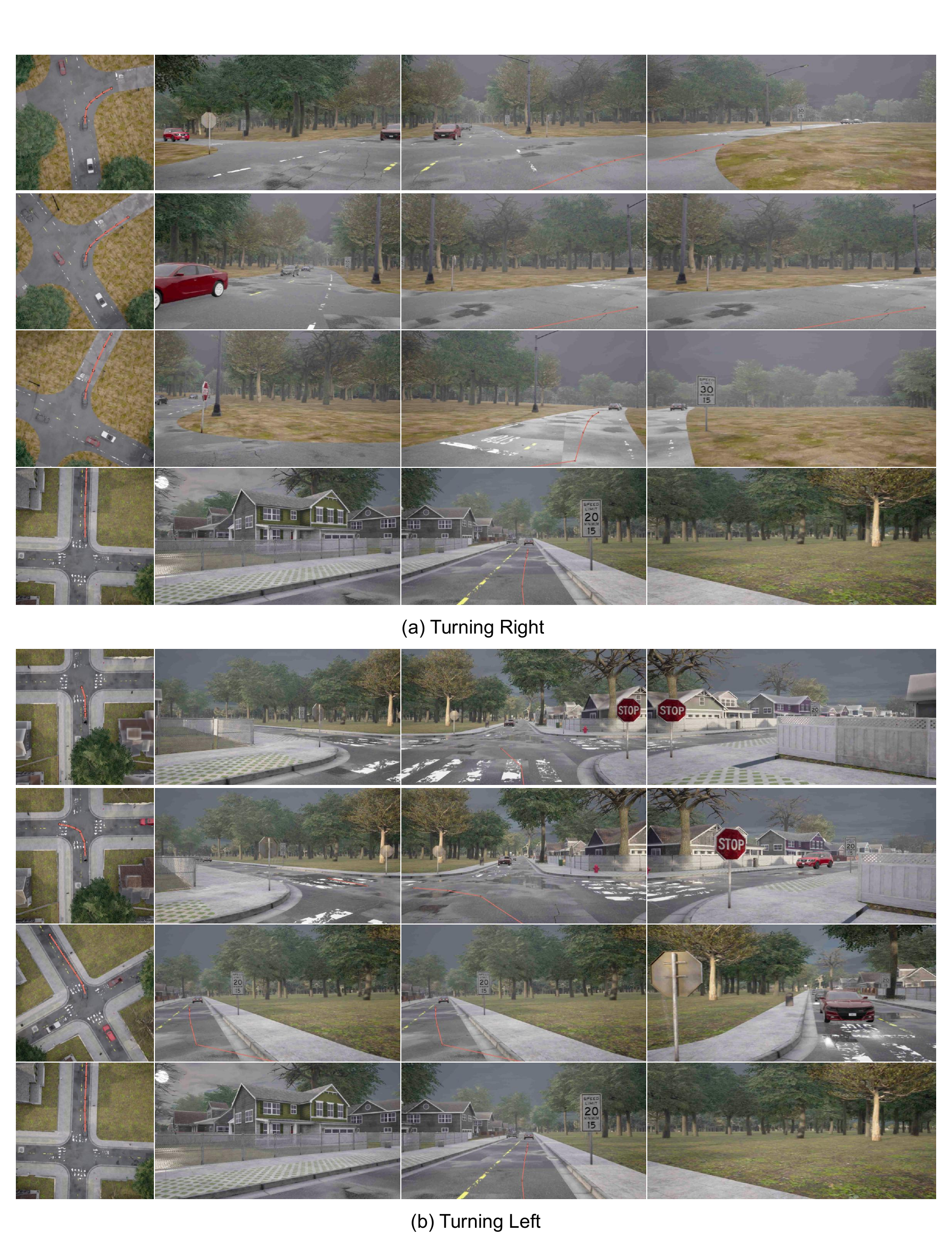}}
    \caption{Qualitative results on the Bench2drive benchmark.}
    \label{fig:nav1}
    \vspace{-5pt}
\end{figure*}

\begin{figure*}[htbp]
\centering
\resizebox{0.97\linewidth}{!}{
\includegraphics[width=1.00\linewidth]{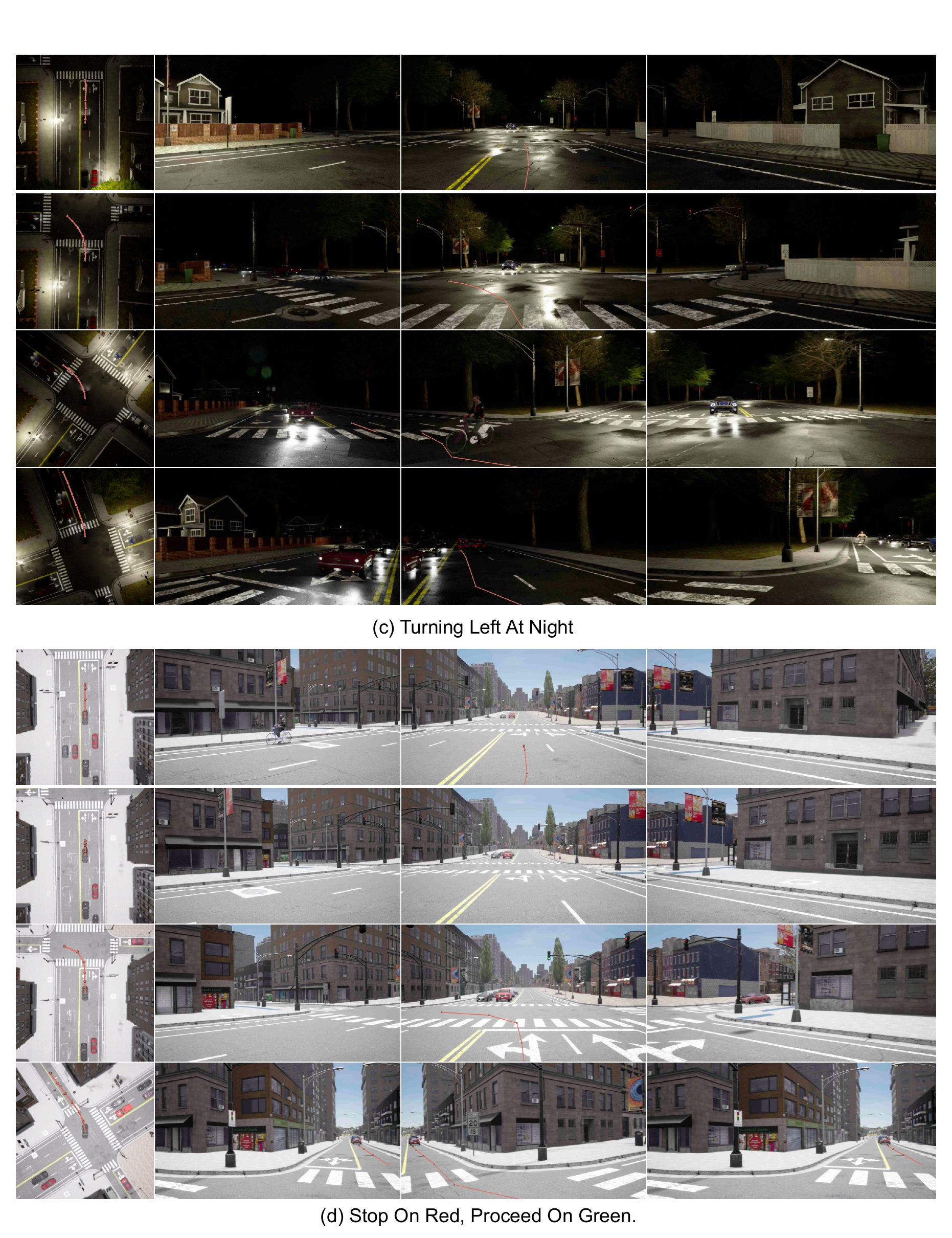}}
    \caption{Qualitative results on the Bench2drive benchmark.}
    \label{fig:nav1}
    \vspace{-5pt}
\end{figure*}

\subsection{Comparison Before and After Reinforcement Learning}
\begin{figure*}[htbp]
\centering
\resizebox{0.97\linewidth}{!}{
\includegraphics[width=1.00\linewidth]{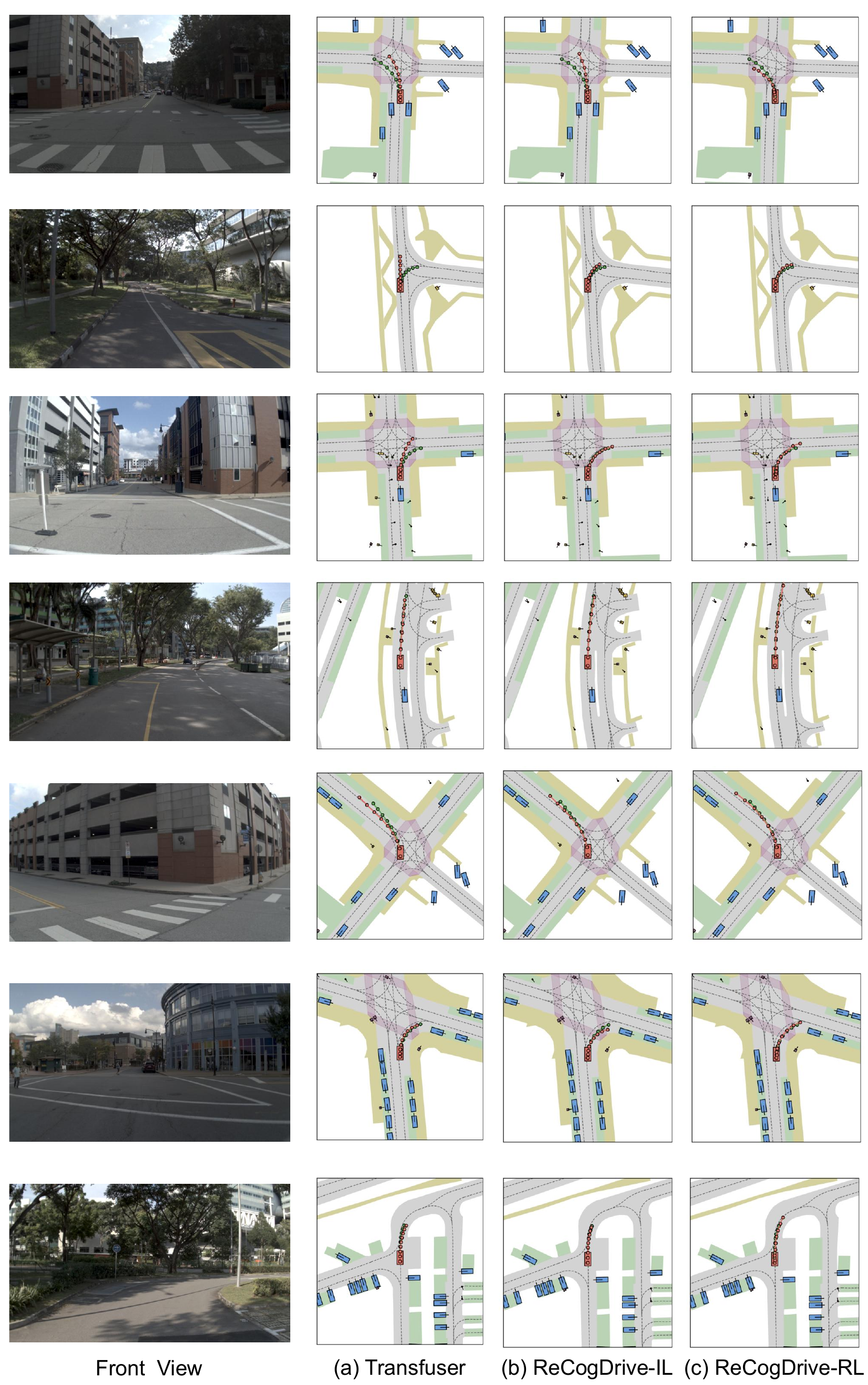}}
    \caption{Qualitative comparisons on the Navtest benchmark.}
    \label{fig:com1}
    \vspace{-5pt}
\end{figure*}

\begin{figure*}[htbp]
\centering
\resizebox{0.97\linewidth}{!}{
\includegraphics[width=1.00\linewidth]{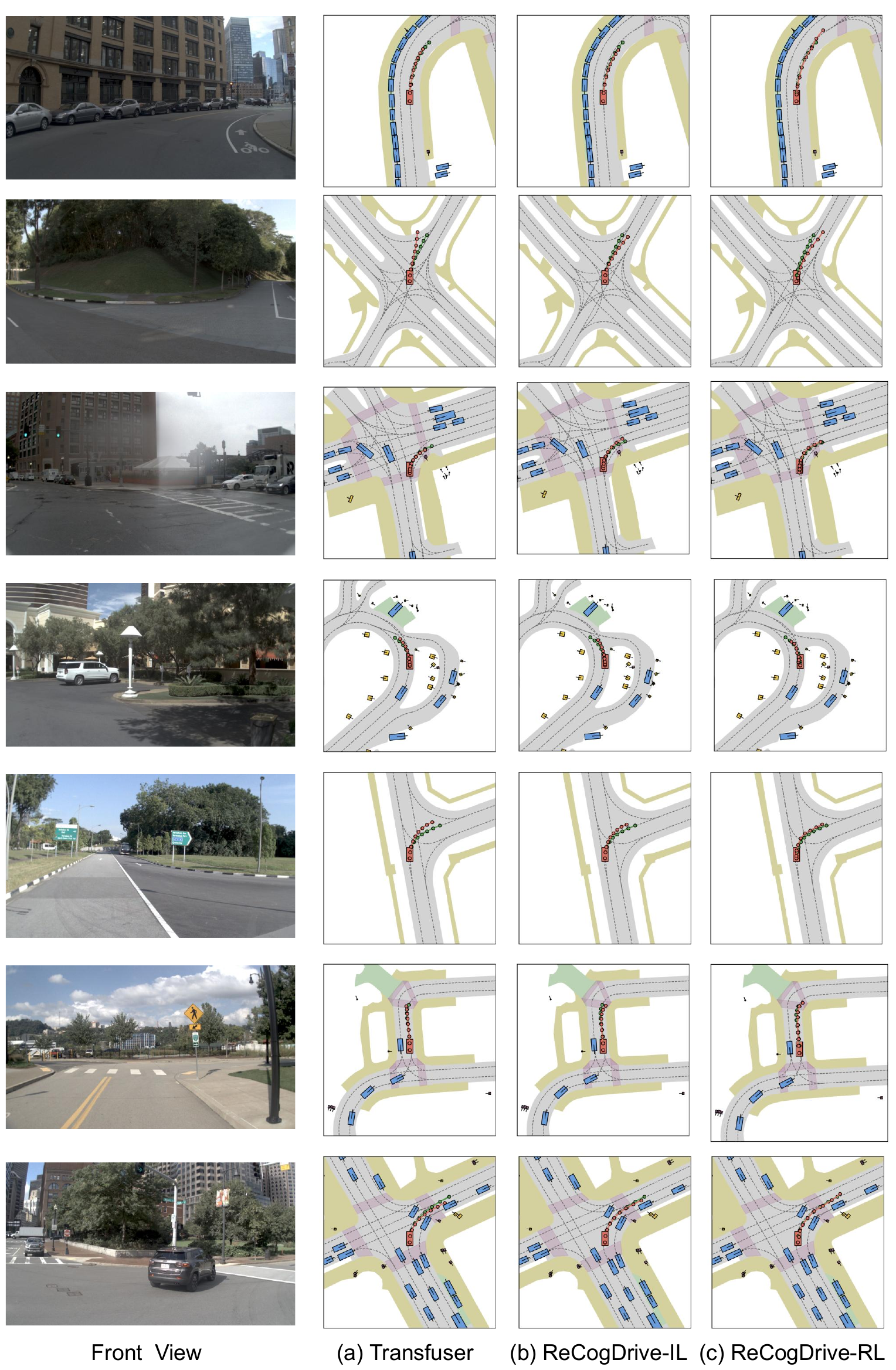}}
    \caption{Qualitative comparisons on the Navtest benchmark.}
    \label{fig:com2}
    \vspace{-5pt}
\end{figure*}

\begin{figure*}[htbp]
\centering
\resizebox{0.97\linewidth}{!}{
\includegraphics[width=1.00\linewidth]{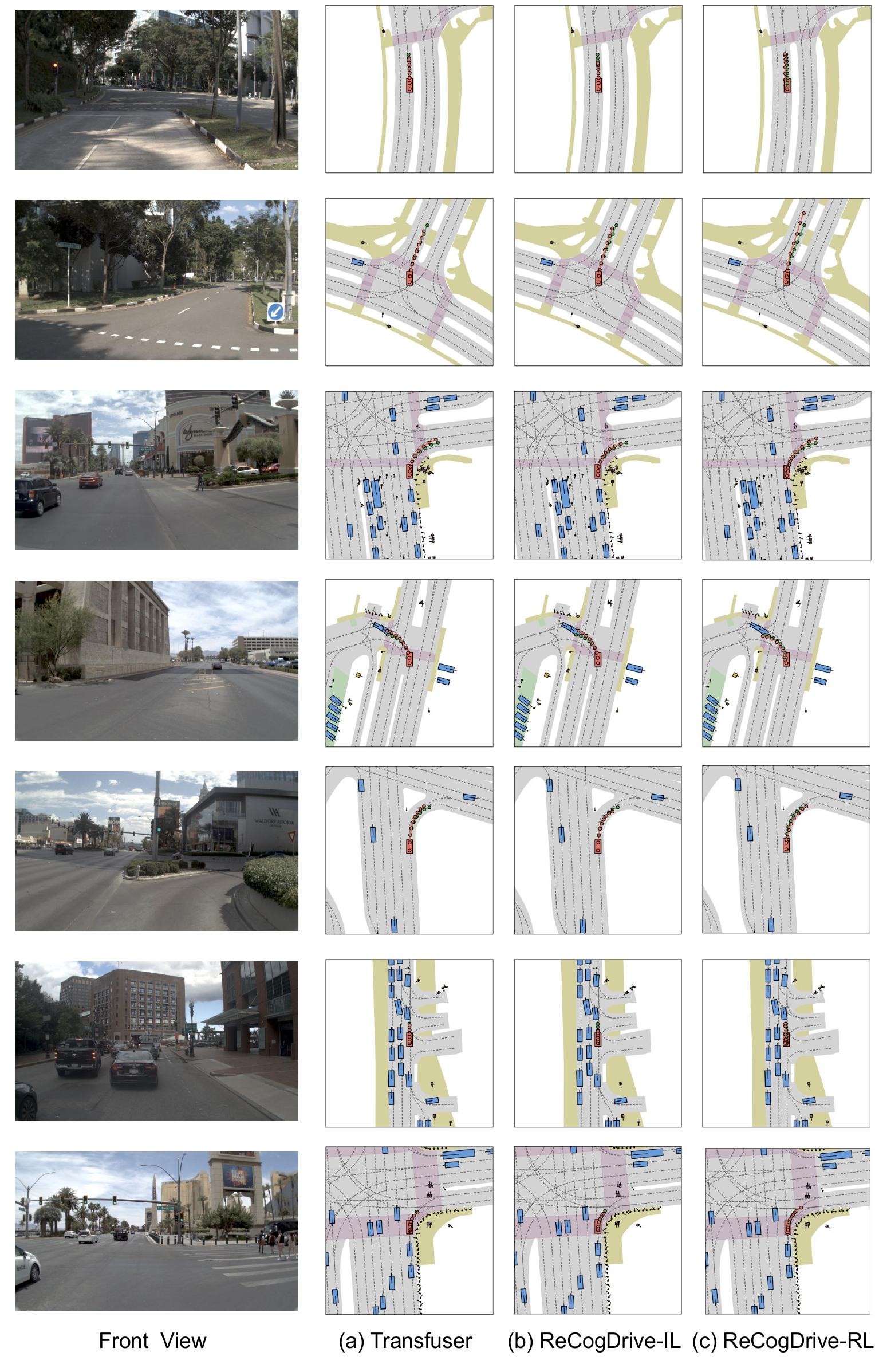}}
    \caption{Qualitative comparisons on the Navtest benchmark.}
    \label{fig:com3}
    \vspace{-5pt}
\end{figure*}

\begin{figure*}[htbp]
\centering
\resizebox{0.97\linewidth}{!}{
\includegraphics[width=1.00\linewidth]{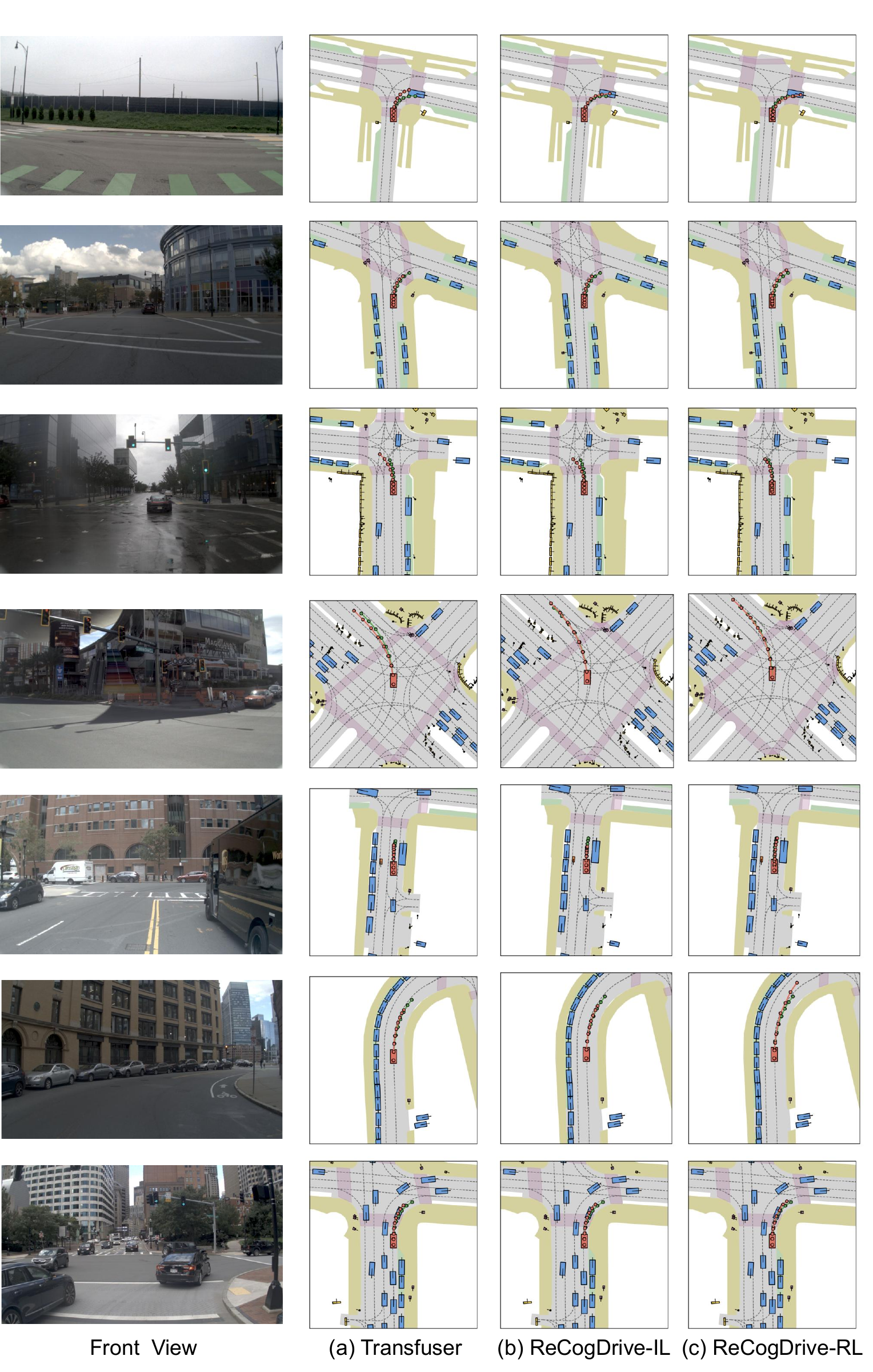}}
    \caption{Qualitative comparisons on the Navtest benchmark.}
    \label{fig:com4}
    \vspace{-5pt}
\end{figure*}

\begin{figure*}[htbp]
\centering
\resizebox{0.97\linewidth}{!}{
\includegraphics[width=1.00\linewidth]{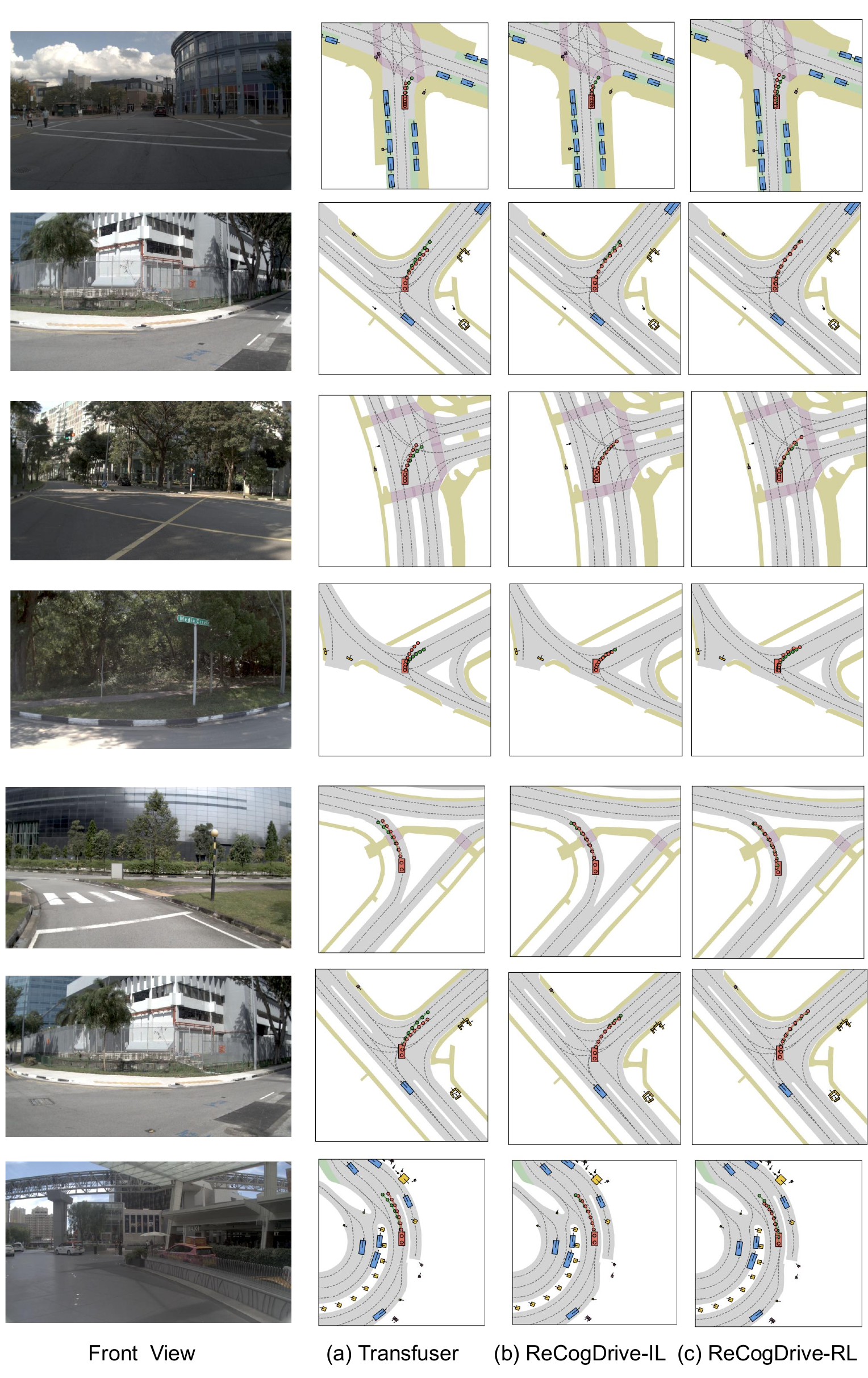}}
    \caption{Qualitative comparisons on the Navtest benchmark.}
    \label{fig:com5}
    \vspace{-5pt}
\end{figure*}

Representative qualitative comparisons are shown in Fig.~\ref{fig:com1}–\ref{fig:com5}, where we contrast Transfuser, imitation learning (IL), and reinforcement learning (RL) variants of \name{}, demonstrating the progressive improvements in decision making, trajectory stability, and safety awareness.

% \begin{figure*}[htbp]
% \centering
% \resizebox{0.97\linewidth}{!}{
% \includegraphics[width=1.00\linewidth]{sec/figures/ICLR-com4.pdf}}
%     \caption{Qualitative comparisons on the Navtest benchmark.}
%     \label{fig:com4}
%     \vspace{-5pt}
% \end{figure*}

\subsection{Failure Cases}

\begin{figure*}[htbp]
\centering
\resizebox{0.97\linewidth}{!}{
\includegraphics[width=1.00\linewidth]{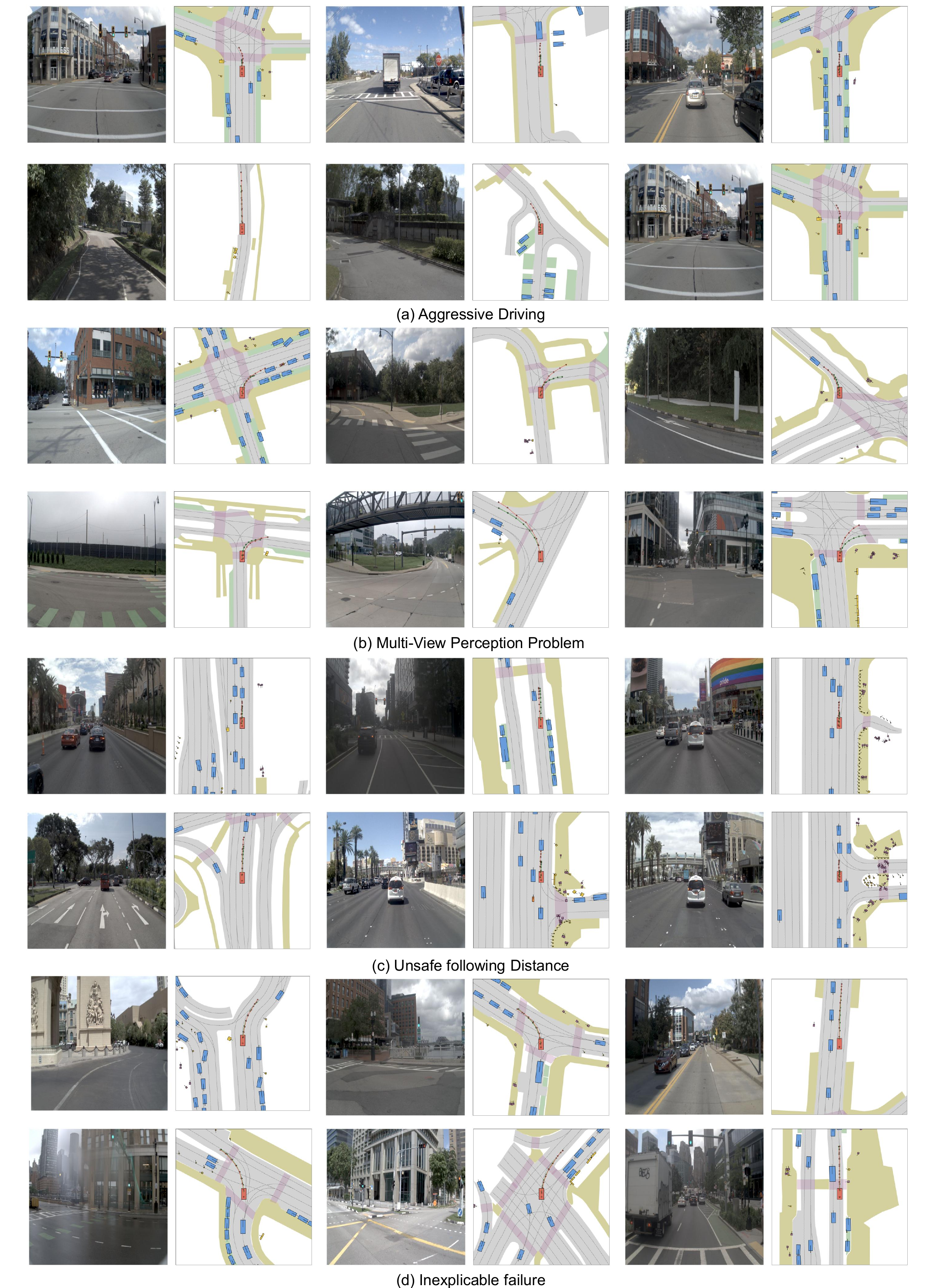}}
    \caption{Representative failure cases of \name{} on the Navsim benchmark, including aggressive driving, multi-view perception problem, unsafe following distances, and inexplicable failures. Zoom in for better view.}
    \label{fig:bad_case}
    \vspace{-5pt}
\end{figure*}

We also present representative failure cases of \name{} on the Navsim benchmark, as shown in Fig.~\ref{fig:bad_case}. These include instances of aggressive driving caused by the model’s tendency to prioritize ego progress, leading to assertive maneuvers. In addition, the relatively weak perception capability of current VLMs means that directly feeding multi-view camera inputs sometimes results in suboptimal performance, particularly in turning scenarios. We also observe unsafe following distances due to insufficient modeling of surrounding vehicle behaviors, reflecting the limitations of VLMs in motion prediction. Finally, we identify cases where predicted trajectories are visually close to the ground truth but still scored as failures, suggesting that the evaluation metrics may suffer from false negatives.

\subsection{Model–Human Dialogue}
\begin{figure*}[htbp]
\centering
\resizebox{0.97\linewidth}{!}{
\includegraphics[width=1.00\linewidth]{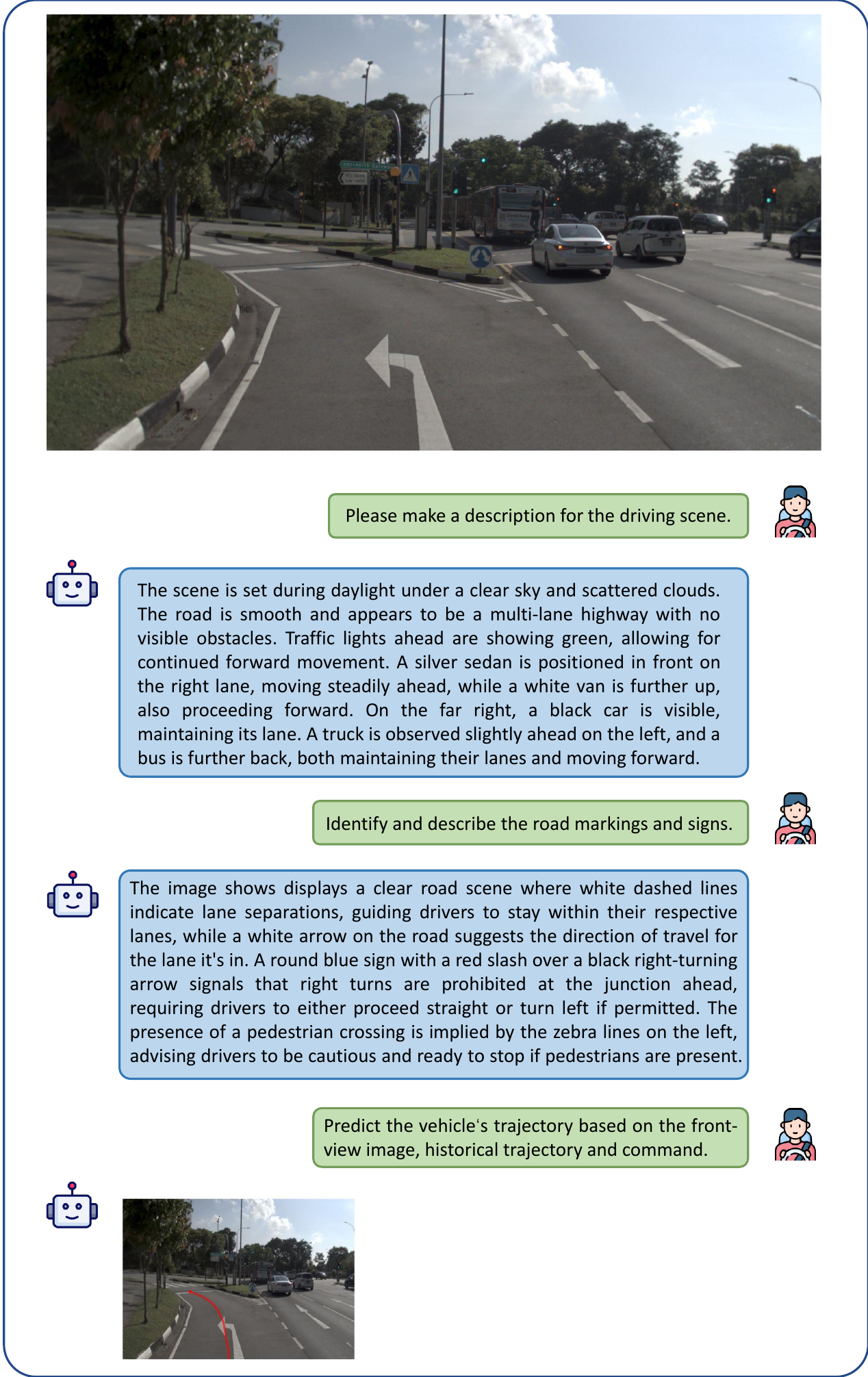}}
    \caption{An example shows \name{}’s capability in road sign recognition and intersection-turn planning: given a front-view image and a user query, \name{} generates a detailed scene description, identifies key traffic signs at the junction, and predicts a safe turning trajectory.}
    \label{fig:qa1}
    \vspace{-5pt}
\end{figure*}

\begin{figure*}[htbp]
\centering
\resizebox{0.97\linewidth}{!}{
\includegraphics[width=1.00\linewidth]{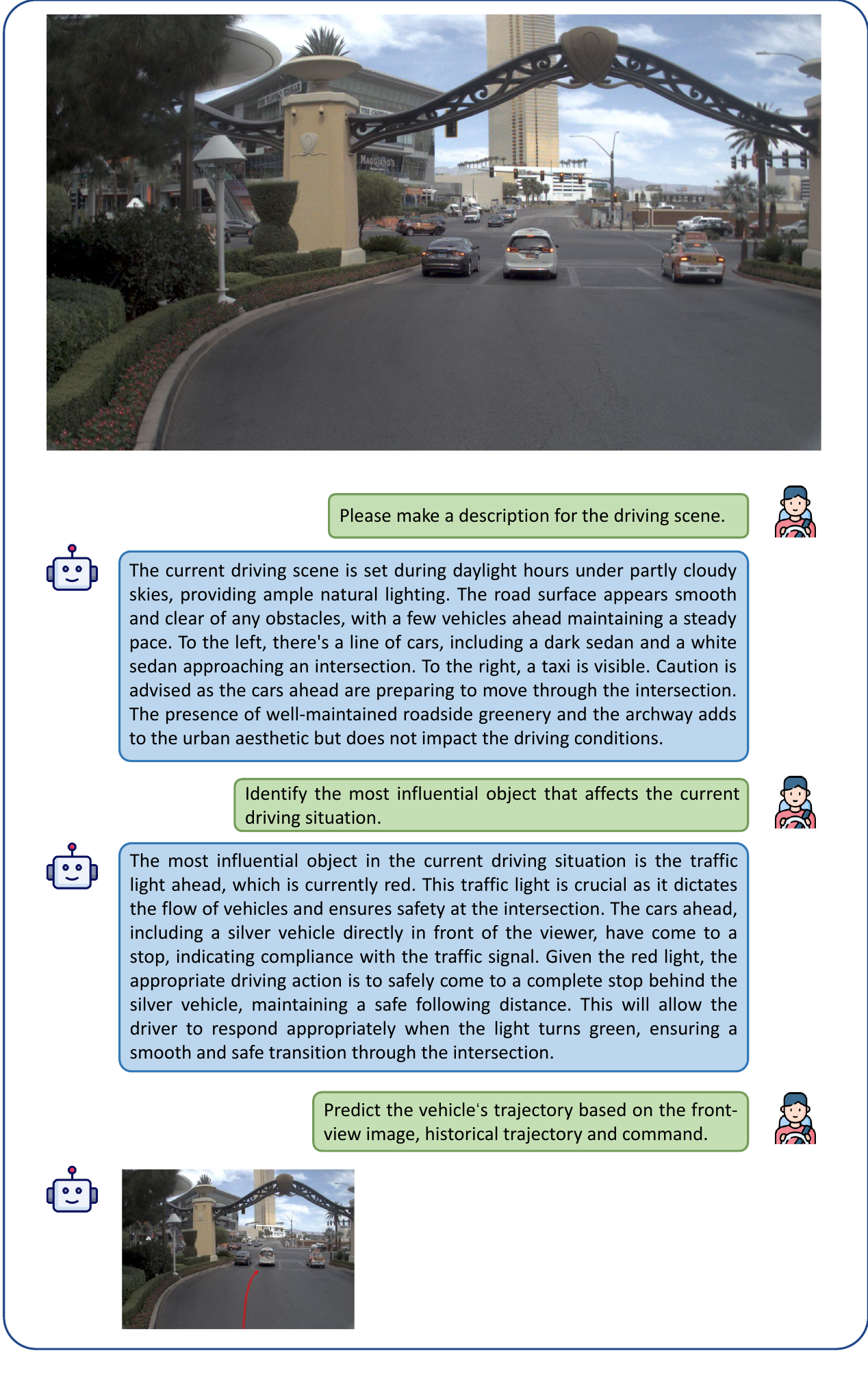}}
    \caption{An example shows \name{}’s capability in scenario understanding and trajectory planning: at a red‐light intersection with braking vehicles ahead, \name{} describes the traffic lights and road conditions, identifies both the traffic light and lead vehicles, and predicts an accurate stop trajectory.}
    \label{fig:qa2}
    \vspace{-5pt}
\end{figure*}

\begin{figure*}[htbp]
\centering
\resizebox{0.97\linewidth}{!}{
\includegraphics[width=1.00\linewidth]{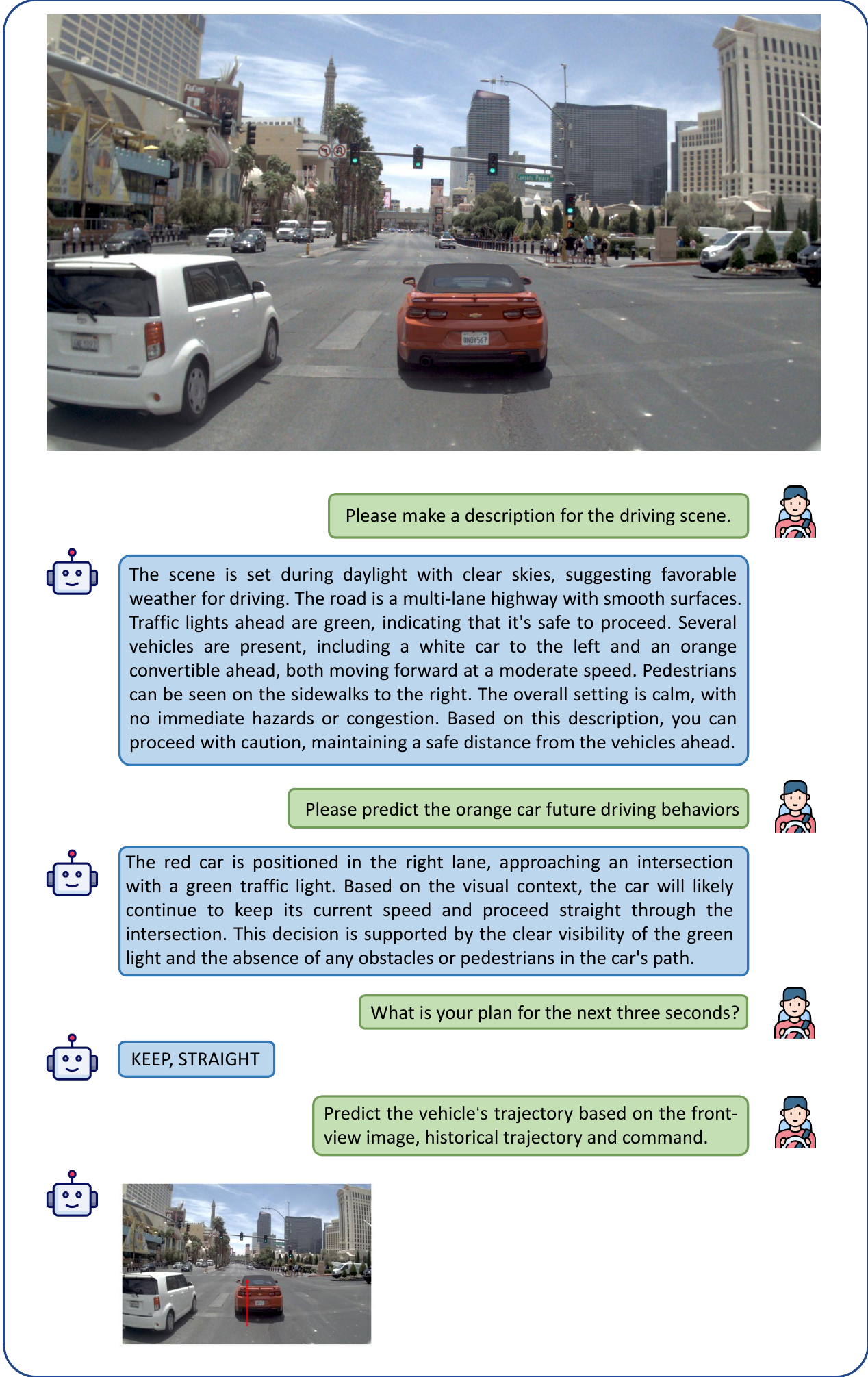}}
    \caption{An example shows \name{}’s capability in behavior prediction and planning: at a green‐light intersection, \name{} describes the road and traffic conditions, identifies the green signal and the lead vehicle’s behavior, generates a high‐level “KEEP,STRAIGHT” command, and predicts the precise continuous trajectory for the ego vehicle.}
    \label{fig:qa3}
    \vspace{-5pt}
\end{figure*}

As shown in Fig.~\ref{fig:qa1}–\ref{fig:qa3}, \name{} engages in interactive dialogue by providing detailed scene descriptions, recognizing traffic signs, reasoning about driving intent, and generating safe turning trajectories.

\end{document}